\pdfoutput=1
\PassOptionsToPackage{table}{xcolor}
\documentclass[11pt]{article}

\usepackage[preprint]{acl}

\usepackage{times}
\usepackage{latexsym}

\usepackage[T1]{fontenc}

\usepackage[utf8]{inputenc}

\usepackage{microtype}
\usepackage{amsmath}
\usepackage{amssymb}
\usepackage[ruled,vlined,linesnumbered]{algorithm2e}
\usepackage{inconsolata}
\usepackage{adjustbox}
\usepackage{bbm}
\usepackage{graphicx}
\usepackage{booktabs} 
\usepackage{multirow}
\usepackage{tabularx}
\usepackage{array}
\usepackage{tcolorbox}
\usepackage{setspace}
\usepackage{enumitem}
\usepackage[table]{xcolor}
\usepackage{arydshln}
\definecolor{groupgray}{gray}{0.93}

\newcommand{\best}[1]{\textbf{#1}}

\newcommand{\second}[1]{\underline{#1}}

\title{PAMT: Process-Aligned Reinforcement Learning for  \\
Multi-Domain Machine Translation}

\author{
Yongshi Ye\textsuperscript{1,3},
Biao Fu\textsuperscript{2,3,}\thanks{\,\,Corresponding authors.},
Chongxuan Huang\textsuperscript{2,3},
Yidong Chen\textsuperscript{2,3},
Xiaodong Shi\textsuperscript{1,2,3,}\footnotemark[1]
\\[0.5em]
\textsuperscript{1}Institute of Artificial Intelligence, Xiamen University \\
\textsuperscript{2}School of Informatics, Xiamen University \\
\textsuperscript{3}Key Laboratory of Digital Protection and Intelligent Processing of Intangible Cultural \\ Heritage of Fujian and Taiwan (Xiamen University), Ministry of Culture and Tourism\\
\texttt{\{yeyongshi,biaofu\}@stu.xmu.edu.cn,mandel@xmu.edu.cn}
}

\begin{document}
\maketitle

\begin{abstract}
Multi-domain machine translation (MDMT) requires more than fluent generation: it demands domain-sensitive translation decisions such as domain disambiguation, terminology control, and stylistic adaptation. Large reasoning models (LRMs) make such decisions explicit through intermediate translation steps, but our analysis across 15 domains and four translation directions shows that this explicit reasoning is double-edged: it improves long-form and high-difficulty translation, yet often drifts in terminology-intensive and stylistically constrained settings. We trace this failure to a credit-assignment bottleneck: existing methods optimize final outputs or coarse trajectories, but cannot identify which translation steps actually help the final translation. To address this, we propose PAMT, a process-aligned training framework that combines cold-start domain-aware Long-CoT supervision with reinforcement learning. PAMT uses sequence-level format and outcome rewards for the final translation, together with a step-level process reward that measures how much each explicit translation step increases the likelihood of the reference translation. Across two backbones, PAMT improves over base models, outperforms MT-specialized baselines on average, and remains competitive with strong LLMs/LRMs across in-domain, OOD, and multilingual settings.
\end{abstract}

\section{Introduction}
\begin{figure*}[t]
    \centering
    \includegraphics[width=\linewidth, trim={0.0cm 5.7cm 0.0cm 0.0cm}, clip]{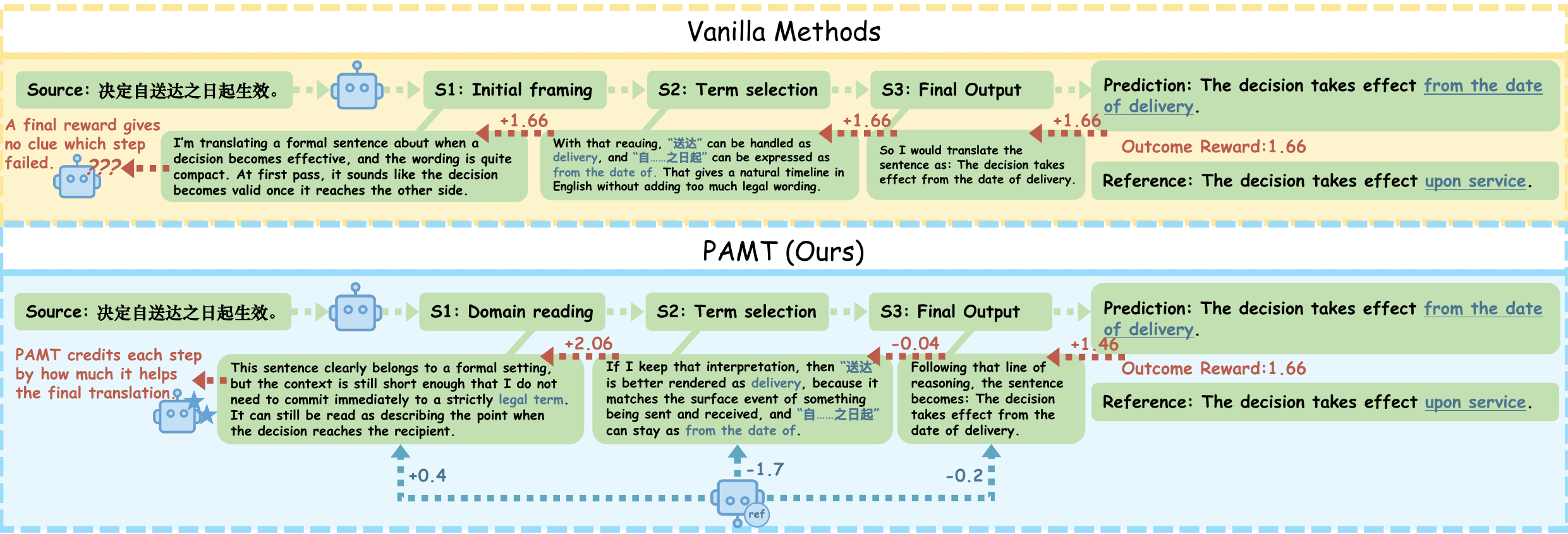}
    \caption{Overview of vanilla reasoning-augmented MT and PAMT. PAMT aligns both the translation process and the final output, addressing the misaligned credit assignment of vanilla methods.}
    \label{fig:damt}
\end{figure*}

Multi-domain machine translation (MDMT) requires more than semantic adequacy. To translate faithfully across domains, a model must make domain-sensitive decisions about ambiguity resolution, terminology, and style~\cite{saunders2022domain,jiang-etal-2020-multi-domain,lai-etal-2022-m4,zheng2024fine,hu-etal-2024-large-language, man2025dmdteval}. However, most translation methods based on large language models (LLMs) still treat translation as a direct sequence generation task~\cite{xu2024a,guo-etal-2024-novel,pang-etal-2025-salute}, without explicitly modeling the intermediate reasoning process. This makes reasoning-oriented translation particularly appealing: explicit intermediate translation steps can, in principle, expose the decision process needed for domain-faithful translation~\cite{he-etal-2024-exploring}. Yet it remains unclear when these decisions help MDMT and when they fail.

To investigate this question, we first conduct a systematic comparison of LLMs and LRMs across 15 domains and four translation directions (Section \ref{sec:preliminary}).
Our analysis reveals a clear split. Explicit reasoning is most helpful on long-context and high-difficulty inputs (Section~\ref{sec:preliminary_help}), where translation benefits from stepwise decomposition and refinement. Yet it is less reliable in terminology-intensive and stylistically constrained settings (Section~\ref{sec:preliminary_hurt}), where seemingly plausible reasoning can drift away from domain-specific conventions. 
These results highlight the need to supervise and control the translation decisions within the translation process.

Recent work has gradually moved from merely exposing the translation process to optimizing it.
Workflow-based methods first make the process explicit, but only through a few coarse stages~\cite{chen-etal-2024-dual,feng-etal-2025-tear,wang-etal-2024-taste}. Chain-of-Thought (CoT)-based methods then provide more detailed process traces~\cite{hu-etal-2024-large-language,he-etal-2024-exploring,wang2025drtdeepreasoningtranslation}, yet still learn them mainly through offline imitation. Reinforcement learning (RL)-based reasoning-augmented methods further optimize translation with reward signals~\cite{he2025r1t1fullyincentivizingtranslation,feng-etal-2025-mt-r1,wang2025deepreasoningtranslationreinforcement,wang2025extransmultilingualdeepreasoning,yang2025ssrzerosimpleselfrewardingreinforcement,li2025tatr1terminologyawaretranslationreinforcement}, but these signals still operate largely on final outputs or whole trajectories. Thus, supervision remains misaligned with translation decisions (Figure~\ref{fig:damt}).
This raises a central question: How can intermediate translation decisions become creditable and optimizable for domain-faithful translation?

We address this problem with \textbf{PAMT}, a \textbf{P}rocess-\textbf{A}ligned training framework for MD\textbf{MT}. 
PAMT first performs cold-start supervised fine-tuning (SFT) on distilled domain-aware Long-CoT data to initialize an explicit translation process. 
It then applies RL to align the translation process with the final translation outcome. Specifically, final translation quality is optimized with sequence-level format and outcome rewards, while the translation process is optimized with a step-level reward defined by how much each explicit translation step increases the likelihood of the reference translation. This process-output alignment breaks the credit-assignment bottleneck in reasoning-augmented MT, allowing the model to optimize not only final translation quality which translation decisions support domain-faithful translation.

\begin{table}[t]
\centering
\small
\resizebox{\linewidth}{!}{%
\begin{tabular}{lcccc}
\toprule
\multirow{2}{*}{Model} & Document-Level & \multicolumn{3}{c}{Sentence-Level} \\

\cmidrule(lr){2-2} \cmidrule(lr){3-5}
& BlonDe & BLEU & COMET & KIWI \\
\midrule
GPT-4o & 35.68 & 36.61 & 89.13 & 84.14 \\
DeepSeek-V3 & 34.61 & \textbf{36.63} & 89.17 & 84.15 \\
Gemini-2.0-Flash & 35.79 & 36.62 & \textbf{89.20} & 83.99 \\
\hdashline
OpenAI-o1 & 29.15 & 35.26 & 88.20 & \textbf{85.63} \\
DeepSeek-R1 & 25.72 & 35.80 & 88.66 & 84.09 \\
OpenAI-o3-mini & 32.98 & 35.38 & 88.96 & 83.91 \\
Gemini-2.0-Flash-Thinking & \textbf{36.17} & 36.56 & 89.09 & 83.90 \\
\bottomrule
\end{tabular}
}
\caption{Document- and sentence-level performance.}

\label{tab:length_analysis}
\end{table}

Our contributions are threefold. 
\textbf{(1)} We systematically evaluate LLMs and LRMs for MDMT across 15 domains and four translation directions, showing that explicit translation reasoning helps in long-form and high-difficulty settings but often fails in terminology-intensive and stylistically constrained ones. 
\textbf{(2)} We identify a key bottleneck in reasoning-augmented MT: supervision granularity is misaligned with decision granularity, making intermediate translation steps hard to credit and optimize, thereby leading to terminology and style drift.
\textbf{(3)} We propose PAMT, a process-aligned training framework that mitigates this bottleneck by aligning translation processes with final translation outcomes, and show on two backbones that it is competitive with SOTA LLMs/LRMs while outperforming MT-specialized baselines across in-domain, out-of-domain, and multilingual settings.

\section{Preliminary Evaluation}
\label{sec:preliminary}

\noindent \textbf{Setup.} 
We compare LLMs and LRMs across 15 domains and four translation directions (En$\leftrightarrow$Zh, En$\leftrightarrow$De), analyzing explicit translation processes along four dimensions: input length, difficulty, Multidimensional Quality Metrics (MQM)-style quality, and terminology accuracy. 
To compare each paradigm at its strongest, we use a best-in-group protocol instead of one-to-one model pairs; setup details are given in Appendix~\ref{appendix:eval_setup}.

\subsection{When Explicit Processes Help} 
\label{sec:preliminary_help}

\noindent \textbf{Document-level Evaluation.} 
\label{sec:length_ass}
On WMT22, we examine whether explicit reasoning becomes more useful with longer context (Table~\ref{tab:length_analysis}). At the sentence level, LRMs do not consistently outperform LLMs: the best LLM achieves the highest BLEU and COMET scores, while the best LRM leads only on COMETKiwi. Still, sentence-level evaluation can miss important errors in cross-sentence consistency and discourse coherence~\citep{laubli-etal-2018-machine,voita-etal-2019-good}. We therefore use BlonDe~\citep{jiang-etal-2022-blonde} to evaluate discourse phenomena such as entities, tense, pronouns, and discourse markers. At the document level, the best LRM obtains the highest BlonDe score (36.17 vs.\ 35.79 for the best LLM), suggesting that explicit reasoning may be more useful for discourse phenomena.

\begin{table*}[t]
\centering
\resizebox{\textwidth}{!}{%
\begin{tabular}{l ccc ccc ccc ccc ccc}
\toprule
\multirow{2}{*}{Model} & \multicolumn{3}{c}{Level 1} & \multicolumn{3}{c}{Level 2} & \multicolumn{3}{c}{Level 3} & \multicolumn{3}{c}{Level 4} & \multicolumn{3}{c}{Level 5} \\
\cmidrule(lr){2-4} \cmidrule(lr){5-7} \cmidrule(lr){8-10} \cmidrule(lr){11-13} \cmidrule(lr){14-16}
 & BLEU & COMET & KIWI & BLEU & COMET & KIWI & BLEU & COMET & KIWI & BLEU & COMET & KIWI & BLEU & COMET & KIWI \\
\midrule
GPT-4o & \textbf{41.96} & \textbf{85.13} & 79.80 & 30.08 & 84.88 & 82.15 & 26.49 & 82.75 & 80.91 & 22.61 & 83.11 & 75.48 & 24.94 & 83.59 & 67.11 \\
DeepSeek-V3 & 40.81 & 85.09 & 79.90 & 27.90 & 84.86 & 82.20 & 25.43 & 82.54 & 80.97 & 22.57 & 82.95 & 75.30 & 25.06 & 83.30 & 67.64 \\
Gemini-2.0-Flash & 41.15 & 84.39 & 79.46 & 29.22 & 84.93 & 81.96 & 27.53 & 82.80 & 80.68 & 24.53 & 82.72 & 74.57 & 28.93 & 83.91 & 66.28 \\
\hdashline
OpenAI-o1 & 40.34 & 84.59 & \textbf{81.10} & 29.67 & 84.88 & \textbf{83.28} & 25.76 & \textbf{82.94} & \textbf{81.94} & 21.99 & \textbf{83.31} & \textbf{76.92} & 22.78 & \textbf{84.02} & \textbf{69.62} \\
DeepSeek-R1 & 41.43 & 84.33 & 79.86 & 27.12 & 84.25 & 81.75 & 24.32 & 82.15 & 80.44 & 18.66 & 82.17 & 75.27 & 19.53 & 82.43 & 68.78 \\
OpenAI-o3-mini & 40.74 & 84.12 & 79.44 & 26.49 & 84.56 & 82.14 & 24.93 & 82.36 & 80.73 & 21.09 & 82.67 & 75.37 & 22.70 & 83.27 & 67.63 \\
Gemini-2.0-Flash-Thinking & 39.87 & 83.84 & 79.36 & \textbf{30.78} & \textbf{84.98} & 82.21 & \textbf{28.15} & 82.68 & 80.62 & \textbf{24.54} & 83.02 & 74.71 & \textbf{28.97} & 83.93 & 66.32 \\
\bottomrule
\end{tabular}%
}
\caption{Performance by translation complexity for LRMs and traditional LLMs.}
\label{tab:metrics_by_level_avg}
\end{table*}

\noindent \textbf{Impact of Translation Difficulty.} 
\label{sec:diff_ass}
To examine how translation difficulty affects model behavior, we use DeepSeek-V3 to assign source sentences from Multi-Domain (De$\rightarrow$En), WMT22 (De$\leftrightarrow$En and Zh$\leftrightarrow$En), and Guofeng WebNovel (Zh$\rightarrow$En) into five difficulty levels (Figure~\ref{fig:difficulty_eval}; Table~\ref{tab:metrics_by_level_avg}). A clear pattern emerges: while traditional LLMs perform competitively on easy inputs, their performance drops sharply as difficulty increases, whereas LRMs surpass them from Level 2 onward and maintain consistent advantages on COMET and COMETKiwi. This gap indicates that performance at higher difficulty is limited not by local generation, but by the translation process itself. As inputs become more complex, success increasingly depends on process-level decisions such as ambiguity resolution, compositional decomposition, and consistency-preserving revision—capabilities that are better supported by explicit reasoning.

\subsection{When Explicit Processes Hurt}
\label{sec:preliminary_hurt}

\begin{table}[t]
\centering
\setlength{\tabcolsep}{3.2pt}
\renewcommand{\arraystretch}{1.05}
\resizebox{\linewidth}{!}{%
\begin{tabular}{lcccccc}
\toprule
\textbf{Model} 
& \textbf{MQM score} $\downarrow$ 
& \textbf{Critical} $\downarrow$ 
& \textbf{Major} $\downarrow$ 
& \textbf{Minor} $\downarrow$ 
& \multicolumn{2}{c}{\textbf{Err. Rate} $\downarrow$} \\
\midrule
GPT-4o & 0.633 & 1.74 & 3.66 & 8.73 & \multicolumn{2}{c}{10.33} \\
DeepSeek-V3 & 0.626 & 1.57 & 4.15 & 9.25 & \multicolumn{2}{c}{10.57} \\
Gemini-2.0-Flash & 0.728 & 2.06 & 4.89 & 9.28 & \multicolumn{2}{c}{10.57} \\
\hdashline
OpenAI-o1 & \textbf{0.502} & \textbf{1.37} & \textbf{2.76} & \textbf{6.86} & \multicolumn{2}{c}{\textbf{8.04}} \\
DeepSeek-R1 & 0.553 & 1.42 & 3.37 & 8.58 & \multicolumn{2}{c}{9.70} \\
OpenAI-o3-mini & 0.880 & 2.18 & 6.19 & 13.77 & \multicolumn{2}{c}{15.31} \\
Gemini-2.0-Flash-Thinking & 0.742 & 1.99 & 5.04 & 9.32 & \multicolumn{2}{c}{10.73} \\
\midrule
\midrule
\textbf{Model} 
& \textbf{Acc. } $\downarrow$ 
& \textbf{Fluency} $\downarrow$ 
& \textbf{Non-trans} $\downarrow$ 
& \textbf{Style} $\downarrow$ 
& \textbf{Term.} $\downarrow$ 
& \textbf{Term Acc.} $\uparrow$ \\
\midrule
GPT-4o & 50.78 & 7.03 & 0.23 & 30.58 & 11.38 & 37.67 \\
DeepSeek-V3 & 50.72 & 6.84 & 0.09 & 30.51 & 11.83 & 37.92 \\
Gemini-2.0-Flash & 52.86 & 6.87 & 0.16 & \textbf{29.43} & \textbf{10.68} & \textbf{38.55} \\
\hdashline
OpenAI-o1 & 50.95 & 7.10 & 0.15 & 29.58 & 12.22 & 37.42 \\
DeepSeek-R1 & 50.50 & 6.68 & \textbf{0.08} & 30.83 & 11.91 & 37.83 \\
OpenAI-o3-mini & \textbf{49.97} & 7.19 & 0.12 & 30.48 & 12.24 & 37.38 \\
Gemini-2.0-Flash-Thinking & 52.54 & \textbf{6.38} & 0.15 & 29.64 & 11.29 & 37.62 \\
\bottomrule
\end{tabular}%
}
\caption{MQM error analysis and terminology accuracy on De$\rightarrow$En. The top panel reports severity-weighted GEMBA-MQM scores, and the bottom panel reports MQM category ratios and terminology accuracy.}
\label{tab:mqm_term}
\end{table}

\noindent \textbf{MQM Analysis.}
\label{sec:mqm_analysis}
We use GEMBA-MQM~\citep{kocmi-federmann-2023-gemba} with DeepSeek-V3 as the automatic MQM annotator to analyze both error severity and error types. Table~\ref{tab:mqm_term} shows that the best LRM has the lowest severity-weighted MQM score and error rate, indicating fewer severe errors overall. We then follow MQM-based expert evaluation practice~\citep{freitag-etal-2021-experts} and inspect fine-grained categories. The strongest LRM still lags behind the strongest LLM on style and terminology errors (29.58/12.22 vs.\ 29.43/10.68), suggesting that explicit reasoning can reduce overall error severity while remaining fragile on domain-sensitive decisions such as register, stylistic convention, and terminology choice.

\noindent \textbf{Terminological Accuracy.}
\label{sec:term_acc_analysis}
We further isolate term-level behavior using the WMT23 Terminology Shared Task~\cite{semenov-etal-2023-findings}, which provides bilingual sentence pairs with aligned source--target terms. For each example, we provide only the source sentence and measure whether the expected target term is correctly produced in the translation (Table~\ref{tab:mqm_term}). The results reinforce the MQM findings: terminology accuracy is substantially less stable than the gains observed on general semantic metrics, and LRMs underperform in settings where translation depends more on faithful lexical realization. This shows that stronger reasoning does not automatically yield stronger term control. Instead, terminology translation is a process-level decision that must remain aligned with domain constraints throughout generation. Together with the MQM analysis, these results point to the same conclusion: the main bottleneck of LRMs in MDMT lies in how intermediate translation steps are supervised.

\section{Related Work}
\label{sec:related_work}
Prior work makes the translation process explicit at different granularities to improve translation quality. Workflow-based methods~\cite{he-etal-2024-exploring,briakou-etal-2024-translating,chen-etal-2024-iterative,chen-etal-2024-dual,wang-etal-2024-taste,ki-carpuat-2024-guiding,feng-etal-2025-tear} decompose translation into fixed stages such as drafting, evaluation, and refinement. Later work introduces domain-aware CoT~\cite{hu-etal-2024-large-language} or multi-agent trajectories~\cite{wang-etal-2025-drt} for SFT, but still learns the process mainly by imitating traces constructed offline. This further motivates RL-based reasoning-augmented MT, where the translation process and final output are optimized through outcome-level feedback from translation metrics~\cite{feng-etal-2025-mt-r1,he2025r1t1fullyincentivizingtranslation} or exemplar-enhanced rewards~\cite{wang2025extransmultilingualdeepreasoning}. DeepTrans~\cite{wang-etal-2026-deeptrans} and TAT-R1~\cite{li2025tatr1terminologyawaretranslationreinforcement} further introduce process- and output-level rewards through external LLM scoring and terminology constraints. Yet these rewards are still assigned over the whole process, making it difficult to identify which intermediate translation step causes an unfaithful translation output.

\section{Method}
To bridge the mismatch between explicit translation steps and optimization, we propose PAMT, a two-stage process-aligned framework: cold-start SFT initializes the explicit process, and RL aligns the process with the final output.

\subsection{Cold Start}
\label{sec:cold_start}
The goal of this stage is to activate the base model's reasoning ability and ensure that it can produce an explicit translation process. We therefore begin with a cold-start SFT stage using a Long-CoT translation dataset distilled from DeepSeek-R1. The dataset contains approximately 7k translation reasoning examples across 10 diverse domains, with around 700 examples per domain. Each example follows the same structured format, \texttt{<think>...</think><answer>...</answer>}, where the model first generates the translation process and then outputs the final translation.

\subsection{RL Stage}
\noindent \textbf{Task Setup.}
During RL, each training example provides a source sentence \(x_i\) for generation and a reference translation \(y_i^*\) only for reward computation. For each \(x_i\), we sample \(G\) policy rollouts, with the \(g\)-th rollout structured as
\begin{equation}
\resizebox{0.88\linewidth}{!}{$\displaystyle
o_{i,g}
=
\texttt{<think>}z_{i,g}\texttt{</think><answer>}y_{i,g}\texttt{</answer>},
$}
\label{eq:response_format}
\end{equation}
where \(z_{i,g}\) denotes the explicit translation process and \(y_{i,g}\) is the final translation.

For step-level alignment, we split the reasoning span by \texttt{\textbackslash n\textbackslash n} into \(K_{i,g}\) reasoning steps:
$z_{i,g}=[z_{i,g}^{(1)}, z_{i,g}^{(2)}, \dots, z_{i,g}^{(K_{i,g})}]$.
Let \(T_{i,g}\) be the response length and \(t_{i,g}^{\mathrm{last}}\) the last valid response token position. Table~\ref{tab:notation} summarizes the notation.

\noindent \textbf{Reward design.}
We use sequence-level rewards to supervise the final response and a step-level reward to assess intermediate reasoning steps.

For response format, we use a simple binary reward: \(r_{i,g}^{\mathrm{fmt}}=1\) if \(\mathrm{Valid}(o_{i,g})=1\), and \(r_{i,g}^{\mathrm{fmt}}=-1\) otherwise, where \(\mathrm{Valid}(\cdot)\) checks whether the output follows the structured format defined in Section~\ref{sec:cold_start}. 
For translation quality, we define the outcome reward as the equal-weighted average of normalized BLEU, COMET, and COMETKiwi. These metrics respectively capture lexical matching, reference-based semantic quality, and reference-free quality estimation, with COMETKiwi included to reduce over-reliance on potentially noisy or domain-specific references. We divide sacreBLEU by 100 and use COMET and COMETKiwi in their native $[0,1]$ ranges:

\begin{equation}
r_{i,g}^{\mathrm{out}}
=
\frac{1}{|\mathcal{Q}|}
\sum_{q\in\mathcal{Q}} \widetilde{q}_{i,g},
\label{eq:outcome_reward}
\end{equation}
where  \(\mathcal{Q}=\{\mathrm{BLEU}, \mathrm{COMET}, \mathrm{COMETKiwi}\}\).

These sequence-level rewards evaluate the format and quality of the final response, but cannot attribute credit to individual reasoning steps.

For process-level alignment, we score each reasoning prefix with a frozen reference model $\pi_{\mathrm{ref}}$. Specifically, for the prefix containing the first \(k\) reasoning steps in rollout \(o_{i,g}\), let \(z_{i,g,\le k}\) denote their concatenation, and define the context as
\begin{equation}
\resizebox{0.88\linewidth}{!}{$\displaystyle
c_{i,g,k}
=
[x_i,\texttt{<think>},z_{i,g,\le k},\texttt{</think><answer>}].
$}
\label{eq:prefix_context}
\end{equation}
We then define the process potential of this prefix as the teacher-forced log-likelihood of the reference translation \(y_i^*\) under \(\pi_{\mathrm{ref}}\):
\begin{equation}
\phi_{i,g,k}
=
\sum_{m=1}^{|y_i^*|}
\log \pi_{\mathrm{ref}}\!\left(y_{i,m}^* \mid c_{i,g,k}, y_{i,<m}^*\right).
\label{eq:process_potential}
\end{equation}

Intuitively, $\phi_{i,g,k}$ measures how well the reasoning prefix up to step $k$ supports the reference translation. If adding step $k$ increases this potential, the step makes the reference more predictable under $\pi_{\mathrm{ref}}$; if it decreases the potential, the step makes the reference less predictable. We capture this effect with the process gain $r_{i,g,k}^{\mathrm{proc}}$:
\begin{equation}
r_{i,g,k}^{\mathrm{proc}}
=
\phi_{i,g,k}-\phi_{i,g,k-1},
\quad
k=1,\dots,K_{i,g},
\label{eq:process_gain}
\end{equation}
where $\phi_{i,g,0}$ denotes the potential of the empty reasoning prefix.

Because optimization is token-level, we uniformly distribute each step gain over that step's tokens. For any token $t$ in $z_{i,g}^{(k)}$, we assign
\begin{equation}
r_{i,g,t}^{\mathrm{proc}}
=
\frac{r_{i,g,k}^{\mathrm{proc}}}{|z_{i,g}^{(k)}|}.
\label{eq:token_process_reward}
\end{equation}
This preserves the total gain of each step while avoiding a bias that would otherwise favor longer steps merely because they contain more tokens.

\noindent \textbf{Credit Assignment.}
We map rewards to token positions based on their granularity. Format and outcome rewards are sequence-level signals placed on the last valid token, while the distributed process reward contributes at the corresponding positions:
\begin{equation}
r_{i,g,t}
=
\mathbf{1}[t=t_{i,g}^{\mathrm{last}}]
\bigl(r_{i,g}^{\mathrm{fmt}}+r_{i,g}^{\mathrm{out}}\bigr)
+
\lambda\, r_{i,g,t}^{\mathrm{proc}},
\label{eq:token_reward}
\end{equation}
where \(\lambda\) controls the weight of the process reward.

For each valid generated token \(t \le t_{i,g}^{\mathrm{last}}\), its return-to-go is
\begin{equation}
{\small
\begin{aligned}
R_{i,g,t}
&=
\sum_{u=t}^{T_{i,g}} r_{i,g,u}
=
r_{i,g}^{\mathrm{fmt}}+r_{i,g}^{\mathrm{out}}
+\lambda\sum_{u=t}^{T_{i,g}} r_{i,g,u}^{\mathrm{proc}}.
\end{aligned}
}
\label{eq:return_to_go}
\end{equation}
This decomposition shows that terminal format and outcome rewards are propagated to all generated tokens through \(R_{i,g,t}\), while process rewards provide dense step-level credit inside the reasoning trace.

\noindent \textbf{Optimization.}
For the $G$ rollouts sampled for the same source sentence $x_i$, let
$R_{i,g}^{\mathrm{traj}}=R_{i,g,1}$ be the total return of rollout $g$.
We compute the group mean $\mu_i$ and standard deviation $\sigma_i$ over
$\{R_{i,g}^{\mathrm{traj}}\}_{g=1}^{G}$, and normalize each token-level return as
\begin{equation}
A_{i,g,t}
=
\frac{R_{i,g,t}-\mu_i}{\sigma_i+\epsilon}.
\label{eq:advantage}
\end{equation}  

We then optimize a GRPO-style clipped surrogate objective with KL regularization:
\begin{equation}
{\small
\begin{aligned}
\mathcal{L}(\theta)
=&\,
-\frac{1}{B G}\sum_{i=1}^{B}\sum_{g=1}^{G}\frac{1}{T_{i,g}}\sum_{t=1}^{T_{i,g}}
\min\!\Biggl(
\rho_{i,g,t}(\theta)A_{i,g,t},\\
&\,
\operatorname{clip}\!\bigl(\rho_{i,g,t}(\theta),1-\epsilon,1+\epsilon\bigr)A_{i,g,t}
\Biggr)\\
&\,
+\beta\,\mathrm{KL}\!\big(\pi_\theta \,\|\, \pi_{\mathrm{ref}}\big).
\end{aligned}
}
\label{eq:pamt_loss}
\end{equation}
where
\begin{equation}
\rho_{i,g,t}(\theta)
=
\frac{
\pi_\theta(o_{i,g,t}\mid x_i,o_{i,g,<t})
}{
\pi_{\theta_{\mathrm{old}}}(o_{i,g,t}\mid x_i,o_{i,g,<t})
}.
\label{eq:importance_ratio}
\end{equation}

Notably, PAMT reuses $\pi_{\mathrm{ref}}$ for both process scoring and KL regularization, requiring no additional scoring model. Algorithm~\ref{alg:pamt} summarizes the RL training procedure, and Appendix~\ref{app:grpo_to_pamt} derives the PAMT objective from standard GRPO.

\section{Experiments}
\label{sec:experiments}

\begin{table*}[t]
\centering
\small
\renewcommand{\arraystretch}{1.08}
\setlength{\tabcolsep}{5pt}
\resizebox{0.95\linewidth}{!}{%
\begin{tabular}{@{}lcccccccccccc@{}}
\toprule
\multirow{2}{*}{Model} & \multicolumn{3}{c}{Laws} & \multicolumn{3}{c}{News} & \multicolumn{3}{c}{Science} & \multicolumn{3}{c}{Subtitles} \\
\cmidrule(lr){2-4} \cmidrule(lr){5-7} \cmidrule(lr){8-10} \cmidrule(lr){11-13}
 & BLEU & COMET & KIWI & BLEU & COMET & KIWI & BLEU & COMET & KIWI & BLEU & COMET & KIWI \\
\midrule
\rowcolor{groupgray}
\multicolumn{13}{c}{\textbf{\textit{Large Language Models}}} \\
DeepSeek-V3 & 59.58 & 89.15 & 84.42 & 35.86 & 86.94 & 85.44 & 33.54 & 88.28 & 85.46 & 23.66 & 82.44 & 82.69 \\
Gemini-2.0-Flash & 56.43 & 88.90 & 84.40 & 36.39 & 86.37 & 85.14 & 33.47 & 87.69 & 85.24 & 24.64 & 81.91 & 82.33 \\
GPT-4o & 48.50 & 88.28 & 84.52 & 34.58 & 86.22 & 85.02 & 32.09 & 87.34 & 84.84 & 24.91 & 81.91 & 81.93 \\
Qwen2.5-7B-Instruct & 43.60 & 86.92 & 83.84 & 31.93 & 85.15 & 83.50 & 30.11 & 86.08 & 83.57 & 25.21 & 79.95 & 81.58 \\
Gemma2-9B-IT & 39.44 & 84.10 & 82.48 & 30.36 & 84.10 & 82.55 & 29.39 & 85.64 & 83.32 & 24.96 & 79.60 & 81.01 \\
\rowcolor{groupgray}
\multicolumn{13}{c}{\textbf{\textit{Large Reasoning Models}}} \\
DeepSeek-R1 & 59.34 & 89.15 & 84.68 & 33.35 & 86.67 & 85.40 & 31.82 & 88.08 & 85.47 & 20.50 & 82.10 & 82.84 \\
Gemini-2.0-Flash-Thinking & 55.75 & 88.61 & 84.27 & 33.90 & 86.02 & 84.79 & 31.98 & 87.47 & 85.06 & 23.83 & 80.90 & 81.22 \\
GPT-5 & 54.07 & 89.34 & 85.43 & 35.05 & 86.65 & 85.52 & 32.21 & 88.06 & 85.47 & 23.23 & 82.32 & 82.68 \\
\rowcolor{groupgray}
\multicolumn{13}{c}{\textbf{\textit{MT-Specialized Models}}} \\
SFT-Parallel-7B & \second{56.64} & \best{88.83} & 84.29 & 28.89 & 85.43 & 83.91 & 29.28 & 86.19 & 83.68 & 26.93 & \second{81.41} & 80.26 \\
ALMA-7B-R & 36.34 & 85.45 & 81.87 & 24.53 & 83.95 & 81.64 & 22.16 & 84.33 & 81.83 & 18.56 & 80.18 & 79.40 \\
ALMA-13B-R & 40.33 & 86.89 & 83.10 & 26.87 & 84.65 & 82.43 & 24.37 & 85.72 & 82.59 & 19.54 & 81.11 & 80.00 \\
TowerInstruct-7B-v0.2 & 50.54 & 88.25 & 82.96 & 30.72 & 84.61 & 82.45 & 27.76 & 85.60 & 82.99 & 22.45 & 80.75 & 80.01 \\
MT-RewardTree & 38.73 & 85.45 & 83.16 & 30.53 & 85.38 & 83.92 & 30.23 & 86.90 & 84.59 & 20.83 & 80.60 & 82.61 \\
CoT-FT-7B & \best{57.05} & \second{88.77} & 84.35 & 30.17 & 85.23 & 83.31 & 28.72 & 86.02 & 83.48 & \best{28.17} & 80.92 & 79.20 \\
MT-R1-Zero-7B & 35.50 & 86.79 & 84.51 & 31.69 & 86.01 & 84.54 & 29.45 & 86.87 & 84.61 & 22.48 & \best{81.73} & 81.70 \\
SSR-X-Zero-7B & 38.88 & 86.44 & 83.44 & 28.14 & 85.87 & 83.98 & 28.54 & 87.11 & 84.65 & 23.18 & 80.50 & 81.99 \\
mExTrans-7B & 38.78 & 87.11 & 84.44 & 25.05 & \second{86.07} & \best{85.34} & 25.74 & \best{87.34} & \best{85.31} & 14.31 & 81.40 & \best{82.85} \\
TAT-R1 & 44.64 & 87.21 & 83.90 & 31.76 & 85.90 & 84.49 & 29.95 & 87.10 & 84.72 & 24.65 & 80.36 & 82.07 \\
\rowcolor{groupgray}
\multicolumn{13}{c}{\textbf{\textit{Our Models}}} \\
PAMT-Qwen2.5-7B-Instruct & 52.11 & 88.58 & \second{84.88} & \best{33.50} & \best{86.11} & \second{84.82} & \best{32.02} & \second{87.25} & \second{84.98} & \second{27.41} & 80.87 & \second{82.62} \\
PAMT-Gemma2-9B-IT & 51.31 & 88.50 & \best{84.90} & \second{32.96} & 86.00 & 84.81 & \second{31.11} & 86.75 & 84.59 & 27.37 & 80.97 & 82.57 \\
\bottomrule
\end{tabular}%
}

\vspace{0.4em}

\setlength{\tabcolsep}{5pt}
\resizebox{0.95\linewidth}{!}{%
\begin{tabular}{@{}lccccccccccccccc@{}}
\toprule
\multirow{2}{*}{Model} & \multicolumn{3}{c}{Literary} & \multicolumn{3}{c}{IT} & \multicolumn{3}{c}{Koran} & \multicolumn{3}{c}{Medical} & \multicolumn{3}{c}{Average} \\
\cmidrule(lr){2-4} \cmidrule(lr){5-7} \cmidrule(lr){8-10} \cmidrule(lr){11-13} \cmidrule(lr){14-16}
 & BLEU & COMET & KIWI & BLEU & COMET & KIWI & BLEU & COMET & KIWI & BLEU & COMET & KIWI & BLEU & COMET & KIWI \\
\midrule
\rowcolor{groupgray}
\multicolumn{16}{c}{\textbf{\textit{Large Language Models}}} \\
DeepSeek-V3 & 16.14 & 77.28 & 77.07 & 38.16 & 83.90 & 78.86 & 17.94 & 74.91 & 80.33 & 41.43 & 84.15 & 81.90 & 33.29 & 83.38 & 82.02 \\
Gemini-2.0-Flash & 18.37 & 77.20 & 76.67 & 37.93 & 83.20 & 78.49 & 19.70 & 74.97 & 79.71 & 44.39 & 84.49 & 81.76 & 33.92 & 83.09 & 81.72 \\
GPT-4o & 17.75 & 77.48 & 77.18 & 37.23 & 83.52 & 78.37 & 17.65 & 75.04 & 80.64 & 41.89 & 84.23 & 81.89 & 31.83 & 83.00 & 81.80 \\
Qwen2.5-7B-Instruct & 14.42 & 76.12 & 75.11 & 30.48 & 80.50 & 77.57 & 15.01 & 72.48 & 79.49 & 35.71 & 81.97 & 81.33 & 28.31 & 81.15 & 80.75 \\
Gemma2-9B-IT & 13.57 & 72.41 & 71.94 & 28.35 & 80.00 & 76.79 & 9.21 & 65.42 & 73.85 & 33.73 & 80.70 & 81.00 & 26.13 & 79.00 & 79.12 \\
\rowcolor{groupgray}
\multicolumn{16}{c}{\textbf{\textit{Large Reasoning Models}}} \\
DeepSeek-R1 & 11.25 & 75.44 & 75.16 & 36.66 & 83.49 & 78.68 & 17.05 & 74.76 & 80.63 & 40.69 & 83.94 & 82.39 & 31.33 & 82.95 & 81.91 \\
Gemini-2.0-Flash-Thinking & 18.01 & 77.06 & 76.63 & 37.26 & 82.94 & 78.40 & 19.57 & 75.03 & 79.79 & 43.36 & 84.14 & 81.70 & 32.96 & 82.77 & 81.48 \\
GPT-5 & 15.01 & 76.72 & 77.08 & 36.92 & 83.58 & 79.22 & 18.83 & 75.49 & 81.80 & 43.00 & 84.29 & 83.13 & 32.29 & 83.31 & 82.54 \\
\rowcolor{groupgray}
\multicolumn{16}{c}{\textbf{\textit{MT-Specialized Models}}} \\
SFT-Parallel-7B & 15.77 & 76.73 & 75.37 & \best{40.64} & \best{84.08} & 79.35 & \best{21.09} & 74.66 & 78.39 & 43.76 & 84.40 & 82.47 & 32.88 & 82.72 & 80.96 \\
ALMA-7B-R & 13.46 & 75.12 & 74.97 & 33.75 & 81.12 & 77.72 & 14.09 & 71.92 & 79.12 & 37.97 & 83.04 & 81.36 & 25.11 & 80.64 & 79.74 \\
ALMA-13B-R & 14.20 & 75.89 & 75.81 & 34.22 & 81.63 & 77.78 & 14.75 & 72.76 & 79.63 & 40.34 & 83.38 & 81.49 & 26.83 & 81.50 & 80.35 \\
TowerInstruct-7B-v0.2 & 15.54 & 75.49 & 74.93 & 38.20 & 83.47 & 78.67 & 10.99 & 69.02 & 70.13 & \best{46.81} & \second{84.58} & 80.80 & 30.38 & 81.47 & 79.12 \\
MT-RewardTree & 14.22 & 76.27 & \second{77.29} & 31.53 & 80.41 & 77.64 & 14.23 & 73.39 & \second{80.92} & 32.30 & 81.59 & 81.21 & 26.57 & 81.25 & 81.42 \\
CoT-FT-7B & 15.31 & 76.45 & 74.96 & \second{40.41} & 83.79 & 79.27 & \second{20.52} & 74.04 & 78.16 & 44.21 & 84.34 & 82.59 & \second{33.07} & 82.44 & 80.67 \\
MT-R1-Zero-7B & 13.75 & 76.93 & 76.72 & 34.51 & 82.71 & 79.59 & 13.05 & 72.84 & 80.66 & 27.28 & 83.31 & 82.69 & 25.96 & 82.15 & 81.88 \\
SSR-X-Zero-7B & 13.66 & 76.82 & 76.14 & 27.01 & 80.08 & 76.57 & 13.97 & 72.86 & 79.75 & 28.86 & 82.12 & 80.98 & 25.28 & 81.47 & 80.94 \\
mExTrans-7B & 10.43 & 76.16 & 76.62 & 25.54 & 78.76 & 77.19 & 11.80 & 73.88 & 80.56 & 25.40 & 81.66 & 81.93 & 22.13 & 81.55 & 81.78 \\
TAT-R1 & 16.63 & 77.50 & 76.80 & 34.43 & 81.57 & 77.67 & 14.79 & 73.29 & 80.61 & 37.39 & 83.16 & 81.71 & 29.28 & 82.01 & 81.50 \\
\rowcolor{groupgray}
\multicolumn{16}{c}{\textbf{\textit{Our Models}}} \\
PAMT-Qwen2.5-7B-Instruct & \second{16.97} & \second{77.96} & 77.26 & 39.11 & 83.75 & \best{80.03} & 18.31 & \second{74.94} & \best{81.10} & 42.12 & 84.13 & \second{82.99} & 32.69 & \second{82.95} & \best{82.33} \\
PAMT-Gemma2-9B-IT & \best{18.06} & \best{78.24} & \best{77.61} & 40.00 & \second{83.85} & \second{79.96} & 19.91 & \best{75.46} & 80.80 & \second{44.65} & \best{84.61} & \best{83.06} & \best{33.17} & \best{83.05} & \second{82.29} \\
\bottomrule
\end{tabular}%
}
\caption{In-domain performance on 8 domains, measured by BLEU, COMET, and COMETKiwi (KIWI). Within MT-Specialized Models and Our Models, the best score is shown in bold and the second-best is underlined.}
\label{tab:indomain-results-styled}
\end{table*}

\begin{table*}[t]
\centering
\small
\renewcommand{\arraystretch}{1.2}
\setlength{\tabcolsep}{3pt}
\resizebox{0.96\linewidth}{!}{%
\begin{tabular}{@{}lcccccccccccccccccc@{}}
\toprule
\multirow{2}{*}{Model} & \multicolumn{3}{c}{Conversation} & \multicolumn{3}{c}{E-commerce} & \multicolumn{3}{c}{Social} & \multicolumn{3}{c}{Culture} & \multicolumn{3}{c}{CommonSense} & \multicolumn{3}{c}{Average} \\
\cmidrule(lr){2-4} \cmidrule(lr){5-7} \cmidrule(lr){8-10} \cmidrule(lr){11-13} \cmidrule(lr){14-16} \cmidrule(lr){17-19}
 & BLEU & COMET & KIWI & BLEU & COMET & KIWI & BLEU & COMET & KIWI & BLEU & COMET & KIWI & BLEU & COMET & KIWI & BLEU & COMET & KIWI \\
\midrule
\rowcolor{groupgray}
\multicolumn{19}{c}{\textbf{\textit{Large Language Models}}} \\
DeepSeek-V3 & 36.76 & 87.03 & 81.48 & 32.29 & 85.53 & 81.19 & 32.26 & 84.59 & 81.44 & 40.27 & 85.46 & 83.21 & 32.55 & 85.36 & 79.98 & 34.83 & 85.59 & 81.46 \\
Gemini-2.0-Flash & 38.05 & 86.90 & 81.37 & 32.41 & 85.49 & 80.87 & 32.98 & 84.28 & 81.04 & 39.11 & 85.02 & 82.94 & 31.60 & 84.71 & 79.38 & 34.83 & 85.28 & 81.12 \\
GPT-4o & 38.02 & 86.79 & 81.43 & 32.88 & 85.57 & 81.04 & 32.75 & 84.47 & 81.08 & 38.65 & 85.24 & 83.14 & 32.54 & 85.26 & 79.87 & 34.97 & 85.47 & 81.31 \\
Qwen2.5-7B-Instruct & 34.14 & 85.35 & 81.02 & 29.22 & 83.89 & 80.41 & 28.12 & 82.62 & 80.25 & 33.60 & 83.04 & 81.76 & 26.37 & 82.42 & 78.52 & 30.29 & 83.46 & 80.39 \\
Gemma2-9B-IT & 33.45 & 83.89 & 79.20 & 27.99 & 82.08 & 78.87 & 26.48 & 80.35 & 78.49 & 32.54 & 81.16 & 79.62 & 28.47 & 83.28 & 79.25 & 29.79 & 82.15 & 79.09 \\
\rowcolor{groupgray}
\multicolumn{19}{c}{\textbf{\textit{Large Reasoning Models}}} \\
DeepSeek-R1 & 33.34 & 86.47 & 81.37 & 27.41 & 85.03 & 80.91 & 27.22 & 83.94 & 81.17 & 35.88 & 85.45 & 83.14 & 28.50 & 84.52 & 79.99 & 30.47 & 85.08 & 81.32 \\
Gemini-2.0-Flash-Thinking & 37.21 & 86.64 & 81.24 & 31.50 & 85.24 & 80.74 & 31.79 & 83.89 & 80.83 & 36.94 & 84.63 & 83.68 & 32.15 & 84.50 & 79.68 & 33.92 & 84.98 & 81.23 \\
GPT-5 & 35.25 & 87.14 & 82.82 & 29.09 & 85.37 & 82.02 & 29.00 & 84.10 & 82.31 & 35.79 & 85.04 & 84.15 & 27.08 & 84.43 & 80.75 & 31.24 & 85.22 & 82.41 \\
\rowcolor{groupgray}
\multicolumn{19}{c}{\textbf{\textit{MT-Specialized Models}}} \\
SFT-Parallel-7B & 32.20 & 84.11 & 80.62 & 26.94 & 82.92 & 80.12 & 25.88 & 81.61 & 80.12 & 31.81 & 82.87 & 81.35 & 22.19 & 81.91 & 78.98 & 27.81 & 82.68 & 80.24 \\
ALMA-7B-R & 29.49 & 84.67 & 79.96 & 25.13 & 82.98 & 79.50 & 25.76 & 82.51 & 79.77 & 33.25 & 83.81 & 82.83 & 23.25 & 82.80 & 80.00 & 27.38 & 83.35 & 80.41 \\
MT-RewardTree & 31.34 & 85.59 & 81.82 & 27.32 & 84.04 & 81.28 & 26.31 & 83.14 & \best{81.80} & 31.30 & 82.79 & 81.69 & 22.59 & 82.86 & 79.94 & 27.77 & 83.68 & 81.31 \\
CoT-FT-7B & 31.58 & 84.09 & 80.63 & 26.71 & 82.95 & 80.11 & 24.98 & 81.37 & 79.83 & 30.41 & 82.46 & 81.10 & 22.38 & 81.97 & 78.88 & 27.21 & 82.57 & 80.11 \\
MT-R1-Zero-7B & 32.38 & 85.88 & 81.50 & 26.98 & 84.25 & 81.25 & 26.39 & 83.28 & 81.40 & 31.86 & 83.99 & 82.85 & 24.50 & 82.94 & 79.53 & 28.42 & 84.07 & 81.31 \\
SSR-X-Zero-7B & 30.65 & 85.32 & 81.13 & 25.64 & 83.94 & 80.79 & 25.47 & 83.19 & 81.26 & 28.70 & 83.25 & 81.11 & 23.93 & 82.82 & 79.80 & 26.88 & 83.70 & 80.82 \\
mExTrans-7B & 24.53 & 84.72 & 81.15 & 20.46 & 83.52 & 81.24 & 19.60 & 82.29 & \second{81.70} & 27.12 & \best{84.30} & \best{83.92} & 18.61 & 81.92 & 79.23 & 22.06 & 83.35 & 81.45 \\
TAT-R1 & 34.18 & 85.85 & 81.20 & 29.77 & 84.50 & 81.05 & 29.37 & 83.51 & 81.27 & \second{35.36} & 84.17 & 83.27 & 28.18 & \second{84.02} & 79.62 & 31.37 & 84.41 & 81.28 \\
\rowcolor{groupgray}
\multicolumn{19}{c}{\textbf{\textit{Our Models}}} \\
PAMT-Qwen2.5-7B-Instruct & \second{34.77} & \second{86.11} & \second{81.88} & \best{31.20} & \second{84.85} & \second{81.60} & \best{31.10} & \second{83.70} & 81.66 & \best{36.15} & \second{84.23} & \second{83.62} & \second{30.36} & \best{84.30} & \second{80.07} & \best{32.72} & \best{84.64} & \second{81.77} \\
PAMT-Gemma2-9B-IT & \best{35.22} & \best{86.13} & \best{82.05} & \second{30.30} & \best{84.93} & \best{81.66} & \second{30.98} & \best{83.73} & 81.60 & 34.53 & 84.02 & 83.59 & \best{31.17} & 83.81 & \best{80.22} & \second{32.44} & \second{84.52} & \best{81.83} \\
\bottomrule
\end{tabular}%
}
\caption{Out-of-domain (OOD) performance on 5 domains, measured by BLEU, COMET, and KIWI. }
\label{tab:ood-results-styled}
\end{table*}

\begin{table*}[t]
\centering
\small
\renewcommand{\arraystretch}{1.2}
\setlength{\tabcolsep}{3pt}
\resizebox{0.96\linewidth}{!}{%
\begin{tabular}{@{}lcccccccccccccccccc@{}}
\toprule
\multirow{2}{*}{Model} & \multicolumn{3}{c}{En$\Rightarrow$Zh} & \multicolumn{3}{c}{Zh$\Rightarrow$En} & \multicolumn{3}{c}{De$\Rightarrow$En} & \multicolumn{3}{c}{En$\Rightarrow$X} & \multicolumn{3}{c}{X$\Rightarrow$En} & \multicolumn{3}{c}{Average} \\
\cmidrule(lr){2-4} \cmidrule(lr){5-7} \cmidrule(lr){8-10} \cmidrule(lr){11-13} \cmidrule(lr){14-16} \cmidrule(lr){17-19}
 & BLEU & COMET & KIWI & BLEU & COMET & KIWI & BLEU & COMET & KIWI & BLEU & COMET & KIWI & BLEU & COMET & KIWI & BLEU & COMET & KIWI \\
\midrule
\rowcolor{groupgray}
\multicolumn{19}{c}{\textbf{\textit{MT-Specialized Models}}} \\
ALMA-7B-R & 26.69 & 84.13 & 81.43 & 13.46 & 75.12 & 74.97 & 28.87 & 79.66 & 80.13 & 2.78 & 58.13 & \second{73.13} & 12.01 & 63.94 & 58.17 & 16.76 & 72.20 & 73.57 \\
Tower-Plus-9B & 37.91 & \best{86.76} & \second{84.88} & 16.56 & 77.88 & 77.78 & \best{36.56} & 82.21 & \best{82.36} & \second{6.07} & \second{63.86} & 69.46 & \second{27.99} & \second{81.38} & \second{79.68} & \second{25.02} & \second{78.42} & \second{78.83} \\
SFT-Parallel-7B & \second{38.58} & 86.34 & 83.40 & 15.77 & 76.73 & 75.37 & \second{35.72} & 81.63 & 81.11 & 1.19 & 52.16 & 43.97 & 18.47 & 75.58 & 73.05 & 21.95 & 74.49 & 71.38 \\
MT-R1-Zero-7B & 32.00 & 86.23 & 84.35 & 13.75 & 76.93 & 76.72 & 25.56 & 80.39 & 81.73 & 5.91 & 58.76 & 58.27 & 21.91 & 77.48 & 75.61 & 19.82 & 75.96 & 75.34 \\
SSR-X-Zero-7B & 31.00 & 85.41 & 82.55 & 19.46 & 80.38 & 78.18 & 26.53 & 81.79 & 81.09 & 6.02 & 58.94 & 56.34 & 20.73 & 78.05 & 75.78 & 20.75 & 76.92 & 74.79 \\
mExTrans-7B & 28.63 & \second{86.44} & \best{85.02} & 10.43 & 76.16 & 76.62 & 22.06 & 79.27 & 80.88 & 5.09 & 61.33 & 61.74 & 15.95 & 76.80 & 75.21 & 16.43 & 76.00 & 75.89 \\
\rowcolor{groupgray}
\multicolumn{19}{c}{\textbf{\textit{Our Models}}} \\
PAMT-Qwen2.5-7B-Instruct & \best{38.66} & 86.05 & 83.75 & \second{24.02} & \best{81.68} & \second{79.01} & 32.50 & \second{83.16} & 82.26 & 5.90 & 58.15 & 54.02 & 22.08 & 76.86 & 75.44 & 24.63 & 77.18 & 74.89 \\
PAMT-Gemma2-9B-IT & 37.11 & 85.87 & 83.68 & \best{25.01} & \second{81.54} & \best{79.20} & 33.23 & \best{83.51} & \second{82.31} & \best{8.63} & \best{66.69} & \best{82.87} & \best{32.91} & \best{84.92} & \best{83.52} & \best{27.38} & \best{80.51} & \best{82.31} \\
\bottomrule
\end{tabular}%
}
\caption{Multilingual performance on 5 language settings, measured by BLEU, COMET, and KIWI.}
\label{tab:multilingual-results-styled}
\end{table*}

\begin{table}[t]
\centering
\small
\setlength{\tabcolsep}{4pt}
\renewcommand{\arraystretch}{1.2}

\resizebox{\linewidth}{!}{%
\begin{tabular}{lcccccc}
\toprule
\multirow{2}{*}{Method}
& \multicolumn{3}{c}{In-Domain}
& \multicolumn{3}{c}{Out-of-Domain} \\
\cmidrule(lr){2-4} \cmidrule(lr){5-7}
& BLEU & COMET & KIWI
& BLEU & COMET & KIWI \\
\midrule
\rowcolor{gray!15}
PAMT                     & 29.83 & 82.10 & 81.31 & 32.72 & 84.64 & 81.77 \\
PAMT w/o credit assign  & 29.33 & 81.99 & 81.04 & 31.62 & 84.12 & 81.15 \\
PAMT w/o process reward & 28.51 & 81.79 & 81.14 & 32.05 & 84.33 & 81.41 \\
PAMT w/o quality reward & 24.94 & 81.01 & 80.55 & 26.50 & 83.42 & 81.14 \\
PAMT w/o RL             & 24.45 & 80.77 & 80.59 & 25.10 & 83.50 & 81.17 \\
\bottomrule
\end{tabular}%
}
\caption{Ablation on in-domain and OOD test sets.}
\label{tab:ablation_pamt}
\end{table}

\subsection{Experimental Settings}

\noindent \textbf{Datasets.}
We use two training sets for the two stages of PAMT: a curated 7K domain-aware Long-CoT dataset for cold-start SFT, and a separate 20K multi-domain parallel dataset for RL. Evaluation covers in-domain, OOD, and multilingual test sets over seen and unseen language pairs. Full details are provided in Appendix~\ref{appendix:eval_dataset} and Appendix~\ref{appendix:train_dataset}.

\noindent \textbf{Metrics.}
We report BLEU, COMET~\cite{rei-etal-2020-comet}, and COMETKiwi~\cite{rei-etal-2022-cometkiwi} for translation quality; metric and implementation details are in Appendices~\ref{appendix:eval_metics} and~\ref{appendix:train_detail}, respectively.

\noindent \textbf{Baselines.}
We compare against three baseline groups.
\textbf{\textit{LLMs}} treat MT mainly as direct sequence generation: DeepSeek-V3~\cite{deepseekai2025deepseekv3technicalreport}, Gemini-2.0-Flash~\cite{googledeepmind2024gemini20}, GPT-4o~\cite{openai2024gpt4ocard}, Gemma2-9B-IT~\cite{gemma_2024}, and Qwen2.5-7B-Instruct~\cite{qwen2.5}.
\textbf{\textit{LRMs}} expose explicit reasoning but lack MT-specific process alignment: DeepSeek-R1~\cite{deepseekai2025deepseekr1}, Gemini-2.0-Flash-Thinking~\cite{googledeepmind2024gemini20flashthinking}, and GPT-5~\cite{openai_gpt5}.
\textbf{\textit{MT-specialized models}} include non-reasoning systems, namely TowerInstruct~\cite{tower_llm_2024}, ALMA-R~\cite{xu2024a,xu2024contrastive}, SFT-Parallel, Tower-Plus-9B~\cite{rei2025towerbridginggeneralitytranslation}, and MT-RewardTree~\cite{feng2025mtrewardtreecomprehensiveframeworkadvancing}, as well as reasoning-augmented methods, namely CoT-FT~\cite{hu-etal-2024-large-language}, MT-R1-Zero~\cite{feng-etal-2025-mt-r1}, mExTrans~\cite{wang2025extransmultilingualdeepreasoning}, SSR-X-Zero~\cite{yang2025ssrzerosimpleselfrewardingreinforcement}, and TAT-R1~\cite{li2025tatr1terminologyawaretranslationreinforcement}.

\subsection{Generalization Across Domains}

A key question for process-aligned MT is whether it learns reusable translation behavior or merely overfits to domain-specific heuristics. As shown in Tables~\ref{tab:indomain-results-styled} and~\ref{tab:ood-results-styled}, PAMT improves over its backbones and achieves the best average performance among MT-specialized systems in both in-domain and OOD settings. In contrast, CoT-FT relies on offline CoT imitation and therefore generalizes less reliably under domain shift, while TAT-R1 improves robustness with terminology-aware rewards but still operates mainly at the output level. As a result, neither method can identify which intermediate translation step leads to correct domain-sensitive choices. By attributing changes in final translation likelihood to individual translation steps, PAMT enables step-level credit assignment, reinforcing the decisions that transfer across domains and suppressing those that cause process drift. 
Expert human evaluation is provided in Appendix~\ref{appendix:human_eval}.

\subsection{Generalization Across Languages}

Table~\ref{tab:multilingual-results-styled} reports the multilingual results, which test whether the model learns language-pair-specific patterns or a translation process that transfers across directions. PAMT achieves the strongest average multilingual performance among MT-specialized baselines and our models, with gains that are especially clear in the transfer-heavy En-X and X-En settings. This pattern suggests that PAMT learns more than language-specific templates. We attribute this advantage to step-level process alignment: sequence-level rewards collapse source interpretation, structural transfer, and target-language realization into a single terminal signal, whereas PAMT credits each explicit translation step by its effect on the reference translation likelihood. 
This allows the model to reinforce transferable intermediate decisions while suppressing brittle language-specific shortcuts, leading to stronger generalization across language directions.

\subsection{Ablation Study}

Table~\ref{tab:ablation_pamt} shows that all components of PAMT contribute to the final gains. Removing credit assignment consistently degrades performance in both in-domain and OOD settings, indicating that step-level process gains must be propagated to the corresponding reasoning tokens to become effective optimization signals. Removing the process reward also leads to clear drops, suggesting that sequence-level outcome supervision alone is insufficient for improving intermediate translation reasoning. The largest degradation comes from removing the quality reward, confirming that explicit supervision on the final translation remains indispensable. Finally, removing RL yields the weakest overall results, showing that supervised initialization alone cannot fully translate explicit reasoning traces into translation gains. Overall, these results validate that PAMT relies on the synergy of quality reward, process reward, and token-level credit assignment.

\subsection{PAMT on Terminology and Style Drift.}
\begin{table}[t]
\centering
\resizebox{\linewidth}{!}{%
\begin{tabular}{l ccc}
\toprule
\textbf{Model} & \textbf{Acc. (\%)} $\downarrow$ & \textbf{Fluency (\%)} $\downarrow$ & \textbf{Non-trans (\%)} $\downarrow$ \\
\midrule
CoT-FT-7B & 18.54 & 19.19 & 13.19 \\
SFT-Parallel-7B & 18.29 & 18.41 & 8.79 \\
TAT-R1 & 18.11 & 19.81 & 46.15 \\
PAMT w/o Process & 14.93 & 15.19 & 12.09 \\
\rowcolor{gray!15} PAMT-Qwen2.5-7B-Instruct & \textbf{14.80} & 14.41 & \textbf{6.59} \\
\rowcolor{gray!15} PAMT-Gemma2-9B-IT & 15.32 & \textbf{12.99} & 13.19 \\
\midrule
\midrule
\textbf{Model} & \textbf{Style (\%)} $\downarrow$ & \textbf{Term. (\%)} $\downarrow$ & \textbf{Term. Acc. (\%)} $\uparrow$ \\
\midrule
CoT-FT-7B & 17.58 & 16.89 & 40.44 \\
SFT-Parallel-7B & 16.76 & 16.82 & 40.03 \\
TAT-R1 & 19.33 & 19.19 & 40.95 \\
PAMT w/o Process & 16.25 & 16.64 & 41.72 \\
\rowcolor{gray!15} PAMT-Qwen2.5-7B-Instruct & 15.93 & 16.20 & 41.88 \\
\rowcolor{gray!15} PAMT-Gemma2-9B-IT & \textbf{14.15} & \textbf{14.26} & \textbf{42.97} \\
\bottomrule
\end{tabular}%
}
\caption{MQM error rates and terminology accuracy; lower MQM and higher Term. Acc. are better.}
\label{tab:mqm_damt}
\end{table}

Table~\ref{tab:mqm_damt} directly evaluates whether process supervision mitigates terminology and style drift. PAMT achieves the lowest error rates overall. Compared with CoT-FT, PAMT-Gemma2-9B-IT reduces style and terminology errors by 3.43 and 2.63 points, showing that exposing the process alone is insufficient. The contrast with TAT-R1 is more striking: despite explicit terminology constraints, it exhibits substantially higher style and terminology errors (19.33/19.19 vs.\ 14.15/14.26) and an extremely high non-translation rate (46.15), indicating that output-level constraints cannot stabilize an unaligned process. The ablation further confirms the role of process reward: removing it increases style and terminology errors (from 14.15/14.26 to 16.25/16.64). Notably, PAMT also achieves higher terminology accuracy than TAT-R1 (42.97 vs.\ 40.95). These results show that terminology and style drift arise from misaligned intermediate translation steps, and that PAMT mitigates this bottleneck through step-level credit assignment. We further provide qualitative case studies in Appendix~\ref{apd:case_study}.

\begin{figure}[t]
    \centering
    \includegraphics[width=\linewidth]{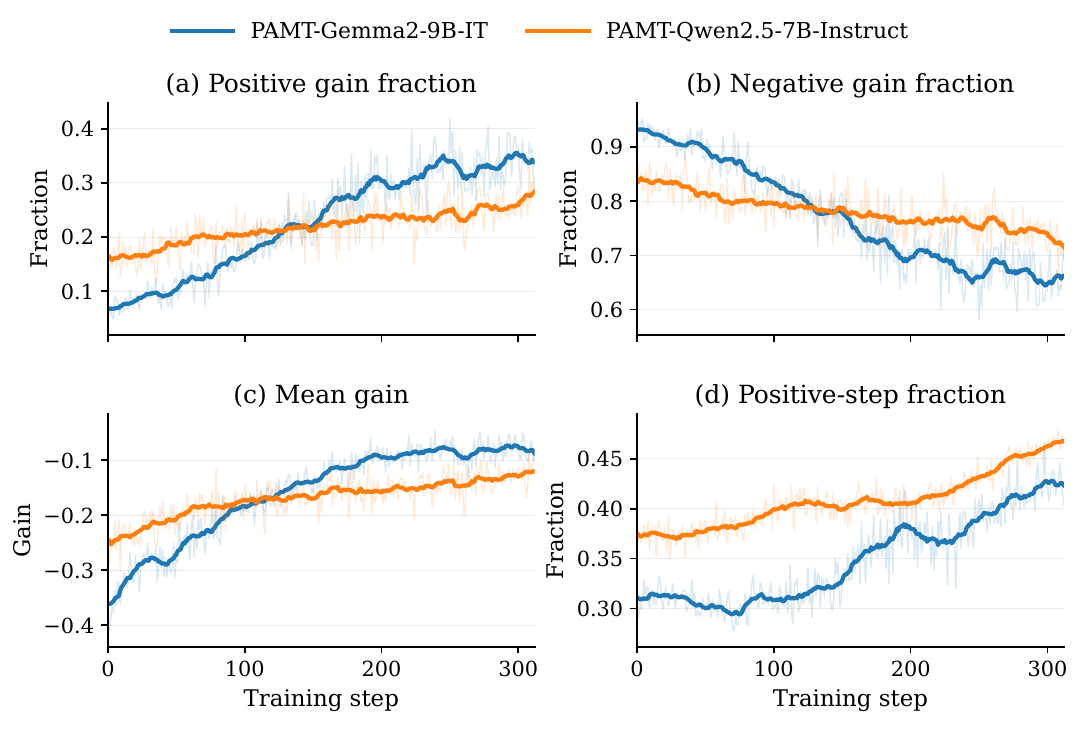}
    \caption{Training dynamics of process rewards}
    \label{fig:training_dynamics}
\end{figure}

\begin{figure}[t]
    \centering
    \includegraphics[width=\linewidth]{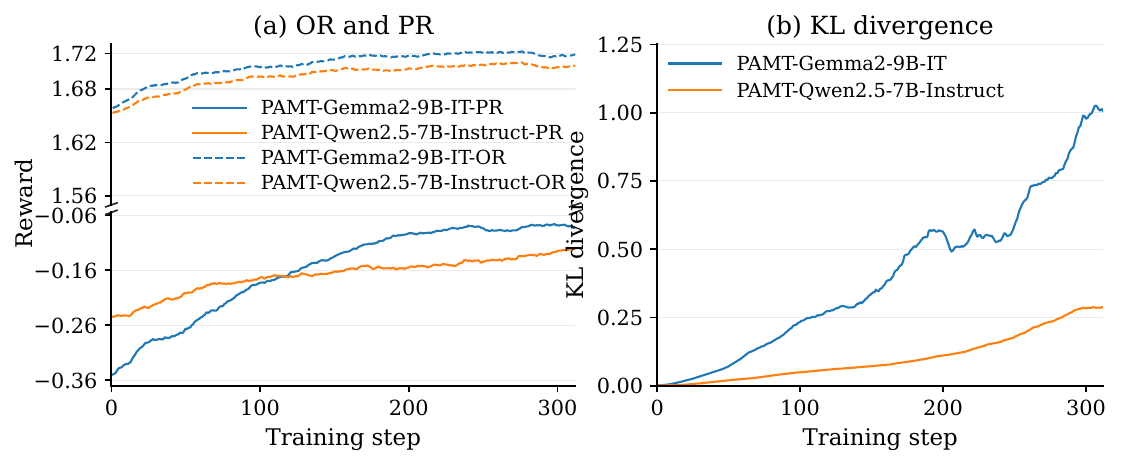}
    \caption{Training dynamics of reward signals and KL.}
    \label{fig:reward_kl_by_step}
\end{figure}

\subsection{Training Dynamics of Process Reward.}
Figure~\ref{fig:training_dynamics} shows that PAMT induces highly consistent process-level learning dynamics across two backbones. As training proceeds, the fraction of positive-gain reasoning steps steadily increases, while the fraction of negative-gain steps decreases. Meanwhile, the mean process gain also improves, indicating that intermediate reasoning steps become progressively less harmful and more supportive of the reference translation. We further observe that the positive-step fraction within each trajectory rises over time, suggesting that the improvement is distributed across the reasoning process rather than concentrated in only a few isolated steps. Overall, these trends provide direct evidence that PAMT effectively optimizes intermediate translation reasoning, rather than only improving the final output.

\subsection{Process Reward Shape Translation Decisions.}
\begin{figure}[t]
    \centering
    \includegraphics[width=\linewidth]{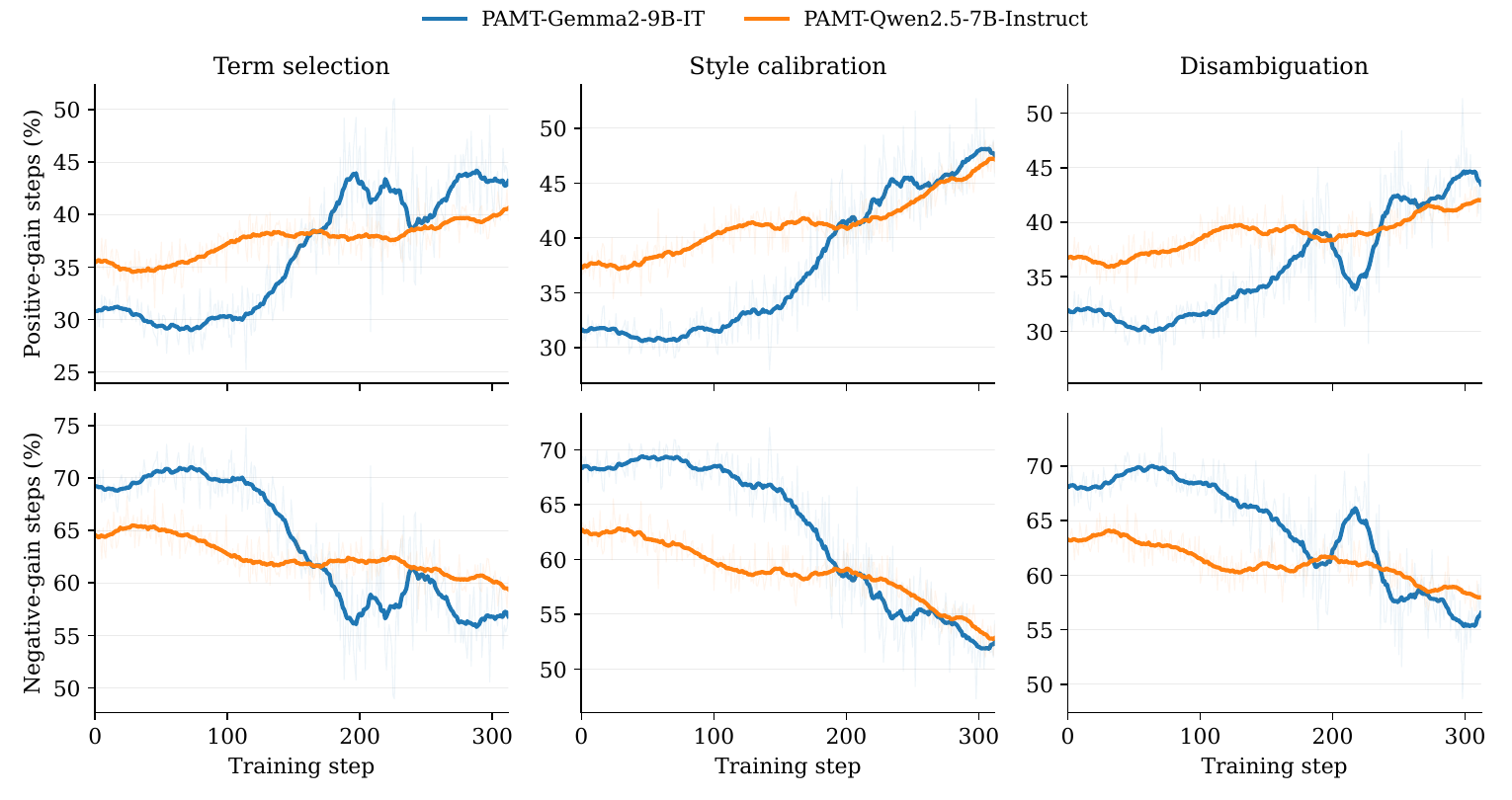}
    \caption{Decision-type percentages in positive- and negative-gain reasoning steps across two PAMT backbones during training.}
    \label{fig:decision_type_gain}
\end{figure}

We further analyze which translation decisions are shaped by the changing process reward observed above. Using keyword and pattern rules, we label each reasoning step as term selection, style calibration, or disambiguation. Figure~\ref{fig:decision_type_gain} shows that, across both backbones, all three decision types become more frequent among positive-gain steps and less frequent among negative-gain steps during training. This indicates that PAMT optimizes domain-sensitive intermediate decisions, rather than only improving final-output rewards.

\subsection{Frozen-Model Overfitting Check.}
As an additional diagnostic, we check whether PAMT overfits to the frozen reference model. Figure~\ref{fig:reward_kl_by_step} shows that outcome reward remains the dominant signal (around 1.7), while process reward is much smaller in scale (about -0.16). The KL from the frozen reference model steadily increases during training. These trends suggest that process reward shapes intermediate steps, while optimization remains driven by final translation quality.

\section{Conclusion}
In this work, we show that explicit translation reasoning is useful but fragile in MDMT: it helps long-form and hard translation, yet can drift on terminology and style when intermediate decisions are not supervised. PAMT addresses this credit-assignment bottleneck by assigning step-level process rewards based on each reasoning step's contribution to the final translation. Across two backbones, PAMT improves its base models, outperforms MT-specialized baselines on average, and remains competitive with strong LLM/LRM systems across in-domain, OOD, and multilingual settings.

\section*{Limitation}
Our study has several limitations. PAMT is designed for explicit translation processes, i.e., intermediate translation steps verbalized in the \texttt{<think>} field, and therefore does not directly model latent internal reasoning that is not exposed in the output trace. In addition, our process reward is reference-based during training, which provides a stable signal for fine-grained credit assignment but assumes access to parallel data. The current implementation also uses simple delimiter-based step segmentation and uniform reward distribution within each step, which is effective in practice but still a coarse approximation of step-level contribution. Finally, process-aware optimization introduces additional training-time cost because multiple step prefixes must be evaluated for each sampled trace, although this does not affect inference. Extending PAMT to weaker-supervision settings, more adaptive process segmentation, and more efficient training remains important future work.

\bibliography{custom,anthology}

\appendix

\begin{table*}[t]
\centering
\small
\setlength{\tabcolsep}{4pt}
\renewcommand{\arraystretch}{1.12}
\begin{tabularx}{\textwidth}{@{}p{0.22\textwidth}X@{}}
\toprule
\textbf{Notation} & \textbf{Meaning} \\
\midrule
\(i\) & Index of a source-reference pair. \\
\(g\) & Rollout index among the \(G\) responses sampled for source \(x_i\). \\
\(t\) & Token position in a generated response. \\
\(k\) & Reasoning-step index in the \texttt{<think>} span. \\
\(m\) & Token position in the reference translation. \\
\(x_i\), \(y_i^*\) & Source sentence and reference translation. The reference is used only for reward computation. \\
\(o_{i,g}\) & The \(g\)-th generated response for source \(x_i\). \\
\(z_{i,g}\), \(y_{i,g}\) & Explicit translation process and final translation in \(o_{i,g}\). \\
\(z_{i,g}^{(k)}\), \(K_{i,g}\) & The \(k\)-th reasoning step and the number of reasoning steps. \\
\(T_{i,g}\), \(t_{i,g}^{\mathrm{last}}\) & Response length and the position of the last valid response token. \\
\(c_{i,g,k}\) & Prefix context containing the first \(k\) reasoning steps. \\
\(\phi_{i,g,k}\) & Process potential, i.e., the teacher-forced log-likelihood of \(y_i^*\) under \(\pi_{\mathrm{ref}}\). \\
\(r_{i,g}^{\mathrm{fmt}}\), \(r_{i,g}^{\mathrm{out}}\) & Sequence-level format and outcome rewards. \\
\(r_{i,g,k}^{\mathrm{proc}}\) & Step-level process gain for reasoning step \(k\). \\
\(r_{i,g,t}^{\mathrm{proc}}\) & Token-level process reward after distributing step gains to tokens. \\
\(r_{i,g,t}\) & Token-level reward combining terminal sequence-level rewards and process reward. \\
\(R_{i,g,t}\) & Return-to-go from token \(t\) onward. \\
\(R_{i,g}^{\mathrm{traj}}\) & Trajectory return, defined as \(R_{i,g,1}\). \\
\(\mu_i\), \(\sigma_i\) & Mean and standard deviation of trajectory returns over the \(G\) rollouts for source \(x_i\). \\
\(A_{i,g,t}\) & Token-level advantage. \\
\(\lambda\), \(\beta\), \(\epsilon\) & Process reward weight, KL coefficient, and clipping/normalization constant. \\
\bottomrule
\end{tabularx}
\caption{List of notation used in PAMT.}
\label{tab:notation}
\end{table*}

\section{Training Setup}
\label{appendix:train_setup}

\subsection{Training Data}
\label{appendix:train_dataset}

\paragraph{Cold-start SFT data.}
The cold-start supervised fine-tuning (SFT) stage uses a curated Long-CoT translation dataset spanning ten domains and three translation directions: De$\rightarrow$En, En$\rightarrow$Zh, and Zh$\rightarrow$En. The dataset contains approximately 7K examples in total. Each example follows the structured format \texttt{<think>...\ </think><answer>...\ </answer>}, where the \texttt{<think>} segment describes an explicit translation process and the \texttt{<answer>} segment provides the final translation. These examples are distilled from strong teacher models to initialize the model's ability to externalize translation reasoning.

\paragraph{RL training data.}
For reinforcement learning (RL), we construct a multilingual multi-domain machine translation training set covering three directions: De$\rightarrow$En, En$\rightarrow$Zh, and Zh$\rightarrow$En. The De$\rightarrow$En portion is drawn from the multi-domain dataset of \citet{aharoni-goldberg-2020-unsupervised}, covering IT, Law, Medical, Koran, and Subtitles. The En$\rightarrow$Zh portion is drawn from UM-Corpus \citep{tian-etal-2014-um}, covering News, Laws, Subtitles, and Science. The Zh$\rightarrow$En Literary domain is taken from the GuoFeng-Webnovel dataset used in the WMT23 and WMT24 literary translation tasks \citep{wang-etal-2023-findings,wang-etal-2024-findings}. From each domain, we randomly sample 2K sentence pairs and retain only examples with a minimum source length of 20 words, or 20 Chinese characters for Chinese inputs, to ensure sufficient room for explicit translation reasoning. The final RL training set contains 20K examples.

\subsection{Implementation Details}
\label{appendix:train_detail}

\paragraph{Cold-start SFT.}
Our SFT implementation is based on LLaMA-Factory\footnote{\url{https://github.com/hiyouga/LLaMA-Factory}}~\cite{zheng-etal-2024-llamafactory}. We adopt Qwen-2.5-7B-Instruct and Gemma2-9B-IT as the base models and fine-tune them on the 7K difficulty-adaptive Long-CoT examples with full-parameter optimization. Training is conducted for 2 epochs on 8 NVIDIA A100 80GB GPUs with a global batch size of 32. We use the AdamW optimizer with a learning rate of $1\mathrm{e}{-5}$, a cosine learning rate scheduler, and a warm-up ratio of 0.1. The maximum input sequence length is set to 4096 tokens. We apply DeepSpeed ZeRO Stage 3 for memory-efficient training.

\paragraph{RL training.}
Our RL implementation is built on \texttt{verl}\footnote{\url{https://github.com/volcengine/verl}}~\cite{sheng2025hybridflow}. We train the model for 2 epochs on 8 NVIDIA A100 80GB GPUs with a global batch size of 128. The rollout number is set to 8, and the rollout temperature is set to 1.0. For PPO updates, the optimization mini-batch size is 16. The learning rate is set to $1\mathrm{e}{-6}$, the KL loss coefficient is $\beta=1\mathrm{e}{-3}$, and the process reward weight is $\lambda=0.1$. The maximum response length is set to 2048 tokens. The entire RL training stage takes about 9 hours.

\paragraph{Inference.}
During inference, we use the \texttt{vLLM}\footnote{\url{https://github.com/vllm-project/vllm}} backend~\citep{woosuk2025vllm} for efficient decoding, with temperature set to 0.0 and a repetition penalty of 1.05.

\section{Evaluation Setup}
\label{appendix:eval_setup}

We use the same evaluation setup for both the preliminary analysis in Section~\ref{sec:preliminary} and the main experiments in Section~\ref{sec:experiments}. This section describes the multi-domain benchmark, the in-domain and out-of-domain test sets, the multilingual evaluation protocol, and the post-training data scale of MT-specialized baselines.

\subsection{Multi-domain Evaluation Benchmark}
\label{appendix:eval_dataset}
\label{appendix:dataset}

\paragraph{Scope and design.}
\label{appendix:eval_dataset_scope}
We build a unified multi-domain benchmark from publicly available corpora with explicit splits. The benchmark serves two purposes. First, it supports the diagnostic analyses in Section~\ref{sec:preliminary} by providing broad coverage across four translation directions: De$\Rightarrow$En, En$\Rightarrow$De, En$\Rightarrow$Zh, and Zh$\Rightarrow$En. Second, it provides the in-domain and out-of-domain evaluation sets used in the main experiments. The benchmark spans both general-purpose and domain-specific text, including biomedical, legal, IT, science, subtitles, conversation, social media, cultural, commonsense, and literary content. It is designed to cover high-resource and low-resource conditions, terminology-intensive domains, context-sensitive inputs, stylistically demanding text, and noisier informal genres, while minimizing potential data leakage and preserving comparability across systems.

\paragraph{Data sources.}
For German$\Leftrightarrow$English, the benchmark covers 11 domains. Five domains---Medical, Law, IT, Koran, and Subtitles---are taken from the publicly available Multi-Domain dataset~\cite{aharoni-goldberg-2020-unsupervised}. The Mixed domain is drawn from the WMT22 General Machine Translation Task~\cite{kocmi-etal-2022-findings}, whose subdomains include News, Social, E-commerce, and Conversation. Biomedical data is collected from the WMT18 and WMT19 Biomedical Machine Translation Tasks~\cite{neves-etal-2018-findings,bawden-etal-2019-findings}.

For Chinese$\Leftrightarrow$English, the benchmark covers 12 domains. The Commonsense and Culture domains are taken from CommonMT~\cite{he-etal-2020-box} and CAMT~\cite{yao-etal-2024-benchmarking}, respectively, while the Literary domain is taken from the WMT23 Literary Machine Translation Task~\cite{wang-etal-2023-findings}. The general-purpose Mixed domain, together with its News, Social, E-commerce, and Conversation subdomains, is again sourced from the WMT22 General Machine Translation Task~\cite{kocmi-etal-2022-findings}. Biomedical data is collected from the WMT18 and WMT19 Biomedical Machine Translation Tasks~\cite{neves-etal-2018-findings,bawden-etal-2019-findings}. In addition, the Laws, News, Science, and Subtitles domains are drawn from UM-Corpus~\cite{tian-etal-2014-um}.

\paragraph{Detailed statistics.}
Tables~\ref{tab:test_zh-en}--\ref{tab:test_en-de} report the number of examples for each domain and direction. No additional filtering or augmentation is applied to these evaluation sets.

\begin{table}[t]
\centering
\small
\begin{tabular}{lc lc}
\toprule
\textbf{Test set} & \textbf{Num} & \textbf{Test set} & \textbf{Num} \\
\midrule
Conversation & 349 & Social & 491 \\
E-commerce & 518 & News (WMT22) & 499 \\
Bio (WMT18) & 212 & Bio (WMT19) & 243 \\
Commonsense & 1200 & Literary & 3038 \\
Terminology & 1638 & Mixed (WMT22) & 1875 \\
\bottomrule
\end{tabular}%
\caption{Test sets and the number of samples for Zh$\Rightarrow$En translation tasks. Bio denotes the Biomedical domain.}
\label{tab:test_zh-en}
\end{table}

\begin{table}[t]
\centering
\small
\begin{tabular}{lc lc}
\toprule
\textbf{Test set} & \textbf{Num} & \textbf{Test set} & \textbf{Num} \\
\midrule
Culture & 778 & Laws & 456 \\
Subtitles & 597 & Science & 503 \\
Conversation & 484 & Social & 511 \\
E-commerce & 530 & News (WMT22) & 511 \\
Bio (WMT18) & 244 & Bio (WMT19) & 224 \\
News (UM) & 1500 & Mixed & 2037 \\
\bottomrule
\end{tabular}%
\caption{Test sets and the number of samples for En$\Rightarrow$Zh translation tasks.}
\label{tab:test_en-zh}
\end{table}

\begin{table}[t]
\centering
\small
\begin{tabular}{lc lc}
\toprule
\textbf{Test set} & \textbf{Num} & \textbf{Test set} & \textbf{Num} \\
\midrule
Bio (WMT18) & 217 & Bio (WMT19) & 373 \\
Social & 515 & E-commerce & 501 \\
News (WMT22) & 506 & Conversation & 462  \\
IT & 2000 & Law & 2000 \\
Koran & 2000 & Subtitles & 2000 \\
Medical & 2000 & Terminology & 2955 \\
Mixed & 1984 & & \\
\bottomrule
\end{tabular}%
\caption{Test sets and the number of samples for De$\Rightarrow$En translation tasks.}
\label{tab:test_de-en}
\end{table}

\begin{table}[t]
\centering
\small
\begin{tabular}{lc lc}
\toprule
\textbf{Test set} & \textbf{Num} & \textbf{Test set} & \textbf{Num} \\
\midrule
Bio (WMT18) & 220 & Bio (WMT19) & 435 \\
News (WMT22) & 511 & E-commerce & 530 \\
Social & 512 & Conversation & 484 \\
Mixed & 2037 & & \\
\bottomrule
\end{tabular}%
\caption{Test sets and the number of samples for En$\Rightarrow$De translation tasks.}
\label{tab:test_en-de}
\end{table}

\subsection{In-Domain Test Sets}
\label{apd:id_data}

For in-domain evaluation, we use the official test sets associated with the corpora used for RL training. Specifically, these sets are drawn from the German-English multi-domain dataset~\citep{aharoni-goldberg-2020-unsupervised}, UM-Corpus~\citep{tian-etal-2014-um}, and the GuoFeng-Webnovel literary dataset~\citep{wang-etal-2023-findings,wang-etal-2024-findings}. For the Literary domain, we merge the \texttt{valid\_1}, \texttt{valid\_2}, \texttt{test\_1}, and \texttt{test\_2} splits into a single evaluation set. Table~\ref{tab:in-domain_test_data} summarizes the in-domain test sets used for the three training directions.

\begin{table}[t]
\centering
\resizebox{\linewidth}{!}{%
\begin{tabular}{lc lc}
\toprule
\textbf{Domain} & \textbf{Num} & \textbf{Domain} & \textbf{Num} \\
\midrule
En$\rightarrow$Zh Laws & 456 & De$\rightarrow$En IT & 2000 \\
En$\rightarrow$Zh Subtitles & 597 & De$\rightarrow$En Koran & 2000 \\ 
En$\rightarrow$Zh Science & 503 & De$\rightarrow$En Medical & 2000 \\
En$\rightarrow$Zh News & 1500 & De$\rightarrow$En Law & 2000 \\
Zh$\rightarrow$En Literary & 3038 & De$\rightarrow$En Subtitles & 2000 \\
\bottomrule
\end{tabular}
}
\caption{In-domain test sets and the number of samples for En$\leftrightarrow$Zh and De$\rightarrow$En translation tasks.}
\label{tab:in-domain_test_data}
\end{table}

\subsection{Out-of-Domain Test Sets}
\label{apd:ood_data}

To evaluate cross-domain generalization, we use public test sets from domains that are not included in the RL training data. Specifically, the Conversation, E-commerce, and Social domains are taken from the WMT22 shared tasks~\cite{kocmi-etal-2022-findings}, the Culture domain is drawn from CAMT~\cite{yao-etal-2024-benchmarking}, and the Commonsense domain comes from CommonMT~\cite{he-etal-2020-box}. These test sets complement the in-domain evaluation by introducing more informal, culturally grounded, and knowledge-sensitive inputs. Table~\ref{tab:out-of_domain_test_data} reports the corresponding statistics.

\begin{table}[t]
\centering
\resizebox{\linewidth}{!}{%
\begin{tabular}{lc lc}
\toprule
\textbf{Domain} & \textbf{Num} & \textbf{Domain} & \textbf{Num} \\
\midrule
En$\rightarrow$Zh Conversation & 484 & En$\rightarrow$Zh Social & 511 \\
En$\rightarrow$Zh Ecommerce & 530 & En$\rightarrow$Zh Culture & 778 \\
Zh$\rightarrow$En CommonSense & 1200 & Zh$\rightarrow$En Conversation & 349 \\
Zh$\rightarrow$En Social & 491 & Zh$\rightarrow$En Ecommerce & 518 \\
De$\rightarrow$En Conversation & 462 & De$\rightarrow$En Social & 515 \\
De$\rightarrow$En Ecommerce & 501 & & \\
\bottomrule
\end{tabular}
}
\caption{Out-of-domain test sets and sample counts for En$\leftrightarrow$Zh and De$\rightarrow$En translation tasks.}
\label{tab:out-of_domain_test_data}
\end{table}

\subsection{Multilingual Evaluation}
\label{apd:multilingual}

For unseen-language evaluation, we use the FLORES+ benchmark~\citep{nllb-24} and construct an \emph{unseen} language set to minimize leakage from languages already covered by baseline post-training data (Table~\ref{tab:multilingual-results-styled}). We first remove all languages appearing in baseline training coverage, including Chinese (zh), English (en), German (de), French (fr), Spanish (es), Portuguese (pt), Italian (it), Russian (ru), Korean (ko), Dutch (nl), Czech (cs), Icelandic (is), Ukrainian (uk), Hindi (hi), Japanese (ja), Polish (pl), Swedish (sv), Hungarian (hu), Romanian (ro), Danish (da), Norwegian (no), and Finnish (fi). We then further restrict the remaining FLORES+ languages to those supported by both COMET and COMETKIWI, so that evaluation is consistent across all unseen directions. After these two filtering steps, the final unseen-language set contains 59 languages, listed in Table~\ref{tab:unseen_lang_list}.

The \emph{seen} languages in our setting are German (de), English (en), and Chinese (zh), which appear in our training data through the German-English multi-domain dataset~\citep{aharoni-goldberg-2020-unsupervised}, UM-Corpus~\citep{tian-etal-2014-um}, and the GuoFeng-Webnovel literary dataset.

\begin{table}[t]
\centering
\resizebox{\linewidth}{!}{%
\begin{tabular}{ll ll ll}
\toprule
ode & Language & Code & Language & Code & Language \\
\midrule
afr & Afrikaans & als & Albanian & amh & Amharic \\
asm & Assamese & bel & Belarusian & ben & Bengali \\
bos & Bosnian & bul & Bulgarian & cat & Catalan \\
cym & Welsh & ekk & Estonian & ell & Greek \\
epo & Esperanto & eus & Basque & fil & Filipino \\
gle & Irish & glg & Galician & guj & Gujarati \\
hau & Hausa & heb & Hebrew & hrv & Croatian \\
hye & Armenian & ind & Indonesian & jav & Javanese \\
kan & Kannada & kat & Georgian & kaz & Kazakh \\
khk & Mongolian & khm & Khmer & kir & Kyrgyz \\
lao & Lao & lit & Lithuanian & lvs & Latvian \\
mal & Malayalam & mar & Marathi & mkd & Macedonian \\
mya & Burmese & npi & Nepali & pan & Punjabi \\
pbt & Pashto & plt & Malagasy & san & Sanskrit \\
sin & Sinhala & slk & Slovak & slv & Slovenian \\
som & Somali & srp & Serbian & sun & Sundanese \\
swh & Swahili & tam & Tamil & tel & Telugu \\
tha & Thai & tur & Turkish & uig & Uyghur \\
urd & Urdu & vie & Vietnamese & xho & Xhosa \\
ydd & Yiddish & zsm & Malay &  &  \\
\bottomrule
\end{tabular}
}
\caption{The 59 unseen languages $\mathcal{L}_{\mathrm{unseen}}$ used for En$\leftrightarrow$X evaluation after filtering FLORES+ by (i) post-training language coverage of evaluated backbones and (ii) COMET/COMETKIWI language support.}
\label{tab:unseen_lang_list}
\end{table}

\subsection{Training Data Scale of MT Baselines}
\label{apd:mt_data_scale}

To facilitate fair comparison, we report the post-training data scale used by each MT-specialized baseline. PAMT and most reasoning-augmented baselines, including MT-R1-Zero-7B, CoT-FT-7B, SFT-Parallel, and mExTrans-7B, are trained on roughly 27K examples. In contrast, several other baselines use substantially larger corpora: TowerInstruct uses 637K examples, Tower-Plus-9B uses 286K, and ALMA-R uses 21K. SSR-X-Zero-7B is trained on a smaller subset of 13K instances.

\subsection{Evaluation Metrics}
\label{appendix:eval_metics}

\paragraph{Automatic Metrics.}
We use three automatic evaluation metrics:
\begin{itemize}
  \item \textbf{BLEU}\footnote{\url{https://github.com/mjpost/sacrebleu}}~\cite{post-2018-call}, which evaluates surface-level n-gram overlap between system output and reference translations.
  \item \textbf{COMET}\footnote{Unbabel/wmt22-comet-da}~\cite{rei-etal-2022-comet}, a reference-based semantic metric trained on human quality judgments.
  \item \textbf{CometKiwi}\footnote{Unbabel/wmt22-cometkiwi-da}~\cite{rei-etal-2022-cometkiwi}, a reference-free version of COMET, useful when reference quality is poor or unavailable.
\end{itemize}

\paragraph{MQM Evaluation Protocol.}
\begin{table*}[t]
\centering
\small
\renewcommand{\arraystretch}{1.15}

\begin{tabularx}{\textwidth}{>{\centering\arraybackslash}p{2.2cm} p{3cm} >{\raggedright\arraybackslash}X}
\toprule
\textbf{Category} & \textbf{Error Type} & \textbf{Description} \\
\midrule

\multirow{8}{*}{Accuracy}
& Mistranslation & Inaccurate translation causing semantic distortion. \\
& Addition & Adding extra information or emotions not in the source. \\
& Under-translation & Failure to fully convey cultural or contextual nuances. \\
& Omission & Unintentional exclusion of content from the source text. \\
& Untranslated & Retaining source text without translation. \\
& Hallucination & Generating content unrelated to the source text. \\
& Off-target Translation & Translation misalignment caused by ambiguous input. \\
& Contradiction & Contradicting itself or the source text. \\

\addlinespace
\midrule

\multirow{5}{*}{Fluency}
& Grammar & Errors in sentence structure or syntax. \\
& Punctuation & Incorrect use of punctuation marks. \\
& Spelling & Misspelling of words. \\
& Semantic Repetition & Unnecessary repetition of words or phrases. \\
& Logical Incoherence & Lack of logical flow or coherence in translation. \\

\addlinespace
\midrule

\multirow{4}{*}{Style}
& Awkward Expression & Stilted or unnatural phrasing in the target language. \\
& Unidiomatic Usage & Literal translation causes unnatural wording or grammar. \\
& Style Inconsistency & Inconsistent stylistic choices within the translation. \\
& Over-localization & Excessive cultural adaptation leading to distortion. \\

\addlinespace
\midrule

\multirow{5}{*}{Terminology}
& Terminology Inconsistency & Inconsistent translation of the same term. \\
& Terminology Misuse & Use of incorrect or inappropriate terms for the domain. \\
& Cross-domain Confusion & Misuse of terms due to domain shifts. \\
& Incorrect Unit Conversion & Errors in unit conversion (e.g., metric to imperial). \\
& Formatting & Errors in domain-specific formats (e.g., legal, medical). \\

\addlinespace
\midrule

\multirow{1}{*}{Others}
& & Any other errors not covered in the above categories. \\

\midrule

\multirow{1}{*}{Source Error}
& & Errors present in the Source text itself. \\

\midrule

\multirow{1}{*}{Non-translation Error}
& & Translation is unassessable and unrelated to the Source. \\

\bottomrule
\end{tabularx}

\caption{\label{tab:mqm-hierarchy} MQM Hierarchy.}

\end{table*}
To evaluate translation quality across domains and systems, we adopt an enhanced MQM hierarchy tailored to the characteristics of LLMs. Based on the official MQM taxonomy, we introduce additional error types frequently observed in LLM outputs—such as hallucination, semantic repetition, and cross-domain confusion. Our final schema spans seven dimensions: Accuracy, Fluency, Style, Terminology, Others, Source Error, and Non-translation Error, covering 25 fine-grained error categories (Table~\ref{tab:mqm-hierarchy}).

For consistent and scalable annotation, we employ DeepSeek-V3 as the automatic scoring model across all evaluations. It is applied uniformly to both traditional and LRMs to ensure fair comparison. The model follows a structured prompt designed to mimic human assessment while enforcing strict formatting and error attribution rules. The full prompt is shown in Figure~\ref{fig:mqm-prompt}.

\subsection{API and Implementation Details}
\label{sec:api}
The OpenAI, DeepSeek, and Gemini models used in this study are accessed via the following APIs: gpt-4o-2024-11-20, o1-2024-12-17, o3-mini-2025-01-31, and gpt-5-2025-08-07 for OpenAI; deepseek-chat-2024-12-26 and deepseek-reasoner for DeepSeek; and gemini-2.0-flash and gemini-2.0-flash-thinking-exp-2025-01-21 for Gemini.

\section{Human--V3 MQM Agreement}
\label{appendix:mqm_agreement}

Automated MQM annotation may introduce bias, but full human MQM annotation is costly at the scale of our evaluation. To validate the reliability of our automatic annotator, we randomly sample 3K examples from the human MQM annotations released by the WMT24 Metrics Shared Task~\citep{freitag-etal-2024-llms} and compare them with DeepSeek-V3 MQM labels under the same error schema. The error-presence agreement reaches 0.8753, supporting the use of DeepSeek-V3 for scalable MQM error analysis. We therefore use automatic MQM as a proxy for aggregate and category-level analysis, while not treating it as a replacement for expert MQM annotation.

\section{Human Evaluation}
\label{appendix:human_eval}

To check whether PAMT's gains are only artifacts of automatic metrics, we conduct a human preference evaluation on 60 examples. Each example is annotated by three expert annotators and compares PAMT against one strong LLM, one strong LRM, and one strong MT baseline. As shown in Table~\ref{tab:human_preference}, PAMT is competitive with GPT-5 and DeepSeek-V3, with more than half of the examples judged as ties, and is clearly preferred over CoT-FT. These results suggest that the improvements are not only due to metric overfitting.

\begin{table}[t]
\centering
\small
\setlength{\tabcolsep}{4pt}
\renewcommand{\arraystretch}{1.1}
\begin{tabularx}{\linewidth}{@{}Xccc@{}}
\toprule
\textbf{Comparison} & \textbf{PAMT win} & \textbf{PAMT lose} & \textbf{Tie} \\
\midrule
PAMT vs. GPT-5 & 15.0 & 26.7 & 58.3 \\
PAMT vs. DeepSeek-V3 & 21.7 & 26.7 & 51.7 \\
PAMT vs. CoT-FT & 61.7 & 11.7 & 26.7 \\
\bottomrule
\end{tabularx}
\caption{Human preference evaluation on 60 examples. Values are percentages.}
\label{tab:human_preference}
\end{table}

\section{Discussion on PRM-Style Step Supervision}
\label{appendix:prm_discussion}

As discussed in Section~\ref{sec:related_work}, recent MT-oriented process reward methods introduce process-level feedback through external LLM scoring or terminology constraints, but they do not isolate the marginal contribution of each explicit translation step. A broader line of PRM-style supervision has been developed for reasoning tasks~\citep{lightman2023letsverifystepstep,wang-etal-2024-math,lai2024stepdpostepwisepreferenceoptimization,luo2024improvemathematicalreasoninglanguage,li-etal-2023-making,pmlr-v202-ni23b}. These methods fit math and code reasoning, where intermediate states often admit relatively clear correctness signals.

Such assumptions are weaker in MT. Translation reasoning is not a chain of uniquely correct proof steps: multiple analyses may support valid translations, and step quality depends on terminology, style, discourse context, and lexical alternatives. PAMT therefore avoids training a separate PRM or collecting human step labels. Instead, it derives step-level credit from the marginal change in reference likelihood under a frozen model, reusing the supervision already available in parallel MT data.

\section{Prompt Templates}
\begin{figure*}[t]
\centering
\begin{tcolorbox}[
colback=gray!10,
colframe=black,
width=\textwidth,
arc=2mm,
boxsep=4pt,
left=6pt,
right=6pt,
top=6pt,
bottom=6pt,
]

\small
\setstretch{1.2}

\noindent
You are a professional translation quality evaluator following the MQM (Multidimensional Quality Metrics) framework.

\medskip
\textbf{Task Instructions:}

\begin{enumerate}[leftmargin=*, itemsep=2pt]
    \item Compare the Prediction against both the Source and the Reference.
    \item Identify up to five of the most serious Errors for each translation sentence, using the MQM error types listed below.
    \item Assign exactly one severity level to each error.
    \item Special handling for two specific error types:
    \begin{itemize}[leftmargin=*, itemsep=1pt]
        \item \textbf{Source Error}: Errors present in the Source text itself.
        \item \textbf{Non-translation Error}: Translation is unassessable and unrelated to the Source.
    \end{itemize}
    \item If no errors are found, return an empty JSON list [].
    \item Only output a valid JSON object. Do not include any additional text, comments, or explanations.
\end{enumerate}

\medskip
\textbf{MQM Error Types (See Table~\ref{tab:mqm-hierarchy} for the complete hierarchy):}

\begin{itemize}[leftmargin=*, itemsep=2pt]
    \item Mistranslation: Incorrect translation that alters or distorts meaning.
    \item Addition: Insertion of information or emotion not present in the source.
    \item Under-translation: Partial omission of relevant or necessary information.
    \item Omission: Complete exclusion of content present in the source.
    \item Untranslated: Source text is copied without being translated.
\end{itemize}

\medskip
\textbf{Severity Levels:}

\begin{itemize}[leftmargin=*, itemsep=2pt]
    \item Minor: Slight impact on readability or style; meaning remains clear.
    \item Major: Significant impact on usability, comprehension, or meaning.
\end{itemize}

\medskip
\textbf{Output Format:}

\medskip
\begin{verbatim}
{
  "errors": [
    {
      "error_type": "Mistranslation",
      "severity": "Major",
      "explanation": "The correct translation should be ... "
    }
  ]
}
\end{verbatim}

\medskip
\textbf{Source:} \texttt{\{source\_text\}} \\
\textbf{Reference:} \texttt{\{reference\_text\}} \\
\textbf{Prediction:} \texttt{\{prediction\_text\}}

\end{tcolorbox}

\caption{Full prompt used to calculate MQM scores with DeepSeek-V3.}
\label{fig:mqm-prompt}

\end{figure*}
\begin{figure*}[t]
\centering
\begin{tcolorbox}[
colback=gray!10,
colframe=black,
width=\textwidth,
arc=2mm,
boxsep=4pt,
left=6pt,
right=6pt,
top=6pt,
bottom=6pt,
]

\small
\setstretch{1.2}

\noindent
Your task is to assess the difficulty of translating a given \texttt{\{src\_lang\}} sentence into \texttt{\{tgt\_lang\}}. Please evaluate the difficulty based on the following criteria and output the result in JSON format, with the key "level":

\medskip
\begin{enumerate}[leftmargin=*, itemsep=2pt]
\item Sentence complexity: Determine if the sentence is a simple sentence, a compound sentence, or includes subordinate clauses and other complex structures.
\item Vocabulary difficulty: Assess whether the sentence contains commonly used words or specialized terms or slang.
\item Grammar differences: Analyze if the sentence's grammatical structure is similar to or differs significantly from \texttt{\{tgt\_lang\}}.
\item Contextual understanding: Consider whether understanding specific cultural contexts or background knowledge is necessary for accurate translation.
\end{enumerate}

\medskip
Provide a difficulty level (1-5), with 1 being the easiest and 5 being the most difficult.

\medskip
And output the difficulty level in the following JSON format:

\medskip
\begin{verbatim}
{
    "level": "difficulty level"
}
\end{verbatim}

\medskip
Here is the \texttt{\{src\_lang\}} sentence: \texttt{\{src\_text\}}

\end{tcolorbox}

\caption{Full prompt used for evaluating translation difficulty with DeepSeek-V3.}
\label{fig:difficulty_eval}

\end{figure*}
\begin{figure*}[t]
\centering
\begin{tcolorbox}[
colback=gray!10,
colframe=black,
width=\textwidth,
arc=2mm,
boxsep=4pt,
left=6pt,
right=6pt,
top=6pt,
bottom=6pt,
]

\small
\setstretch{1.2}

\noindent
Translate the following \texttt{\{src\_lang\}} text into \texttt{\{tgt\_lang\}} while maintaining the domain style of the source text.

\medskip
\textbf{Source:} \texttt{\{src\_text\}}

\end{tcolorbox}

\caption{Full prompt used for cold-start data distillation.}
\label{fig:sft_prompt}

\end{figure*}
\begin{figure*}[t]
\centering
\begin{tcolorbox}[
colback=gray!10,
colframe=black,
width=\textwidth,
arc=2mm,
boxsep=4pt,
left=6pt,
right=6pt,
top=6pt,
bottom=6pt,
]

\small
\setstretch{1.2}

\noindent
Translate the following \texttt{\{src\_lang\}} text into \texttt{\{tgt\_lang\}}.

\medskip
\textbf{Source:} \texttt{\{src\_text\}}

\end{tcolorbox}

\caption{Full prompt used for SFT and RL stage.}
\label{fig:rl_prompt}

\end{figure*}
This section presents the prompts used in our experiments, including those for MQM-style evaluation (Figure~\ref{fig:mqm-prompt}), difficulty assessment (Figure~\ref{fig:difficulty_eval}), cold-start Long-CoT data distillation (Figure~\ref{fig:sft_prompt}), and model training in both SFT and RL stages (Figure~\ref{fig:rl_prompt}).

\section{Case Study: Resolving the Credit Assignment Bottleneck}
\label{apd:case_study}
\begin{figure*}[t]
\centering
\begin{tcolorbox}[
    colback=gray!5!white,
    colframe=black!70!white,
    title=\textbf{Case Study: Trajectory of Process Potential in PAMT (Law Domain)},
    fonttitle=\bfseries\small,
    boxrule=0.8pt,
    arc=3mm,
    left=5pt, right=5pt, top=5pt, bottom=5pt
]
\small

\textbf{Language Pair:} German $\rightarrow$ English (de-en) \hfill
\textbf{Scores:} Format: 1.0000 \,|\, Answer: 2.4624 \,|\, Total: 3.4624 \\
\rule{\linewidth}{0.4pt}
\vspace{2pt}

\textbf{Source Input:} \\
- "Übergangszeitraum": der Zeitraum zwischen dem 1. Januar 1999 und dem 31, Dezember 2001. \\
\vspace{2pt}

\textbf{Reference:} \\
- "transitional period" shall mean the period between 1 January 1999 and 31 December 2001. \\
\vspace{2pt}

\textbf{Prediction:} \\
- "Transitional period": the period between 1 January 1999 and 31 December 2001. \\
\rule{\linewidth}{0.4pt}
\vspace{2pt}

\textbf{Step-by-Step Reasoning Analysis:} \hfill \textit{Initial} $\phi = -30.7478$ \\

\begin{itemize}
    \setlength{\itemsep}{2pt}
    \setlength{\parskip}{0pt}
    
    \item \textbf{Step 1:} Okay, let's tackle this translation. The original German sentence is about a term: "Übergangszeitraum": der Zeitraum zwischen dem 1. Januar 1999 und dem 31. Dezember 2001. \\
    \textit{Metrics:} $\phi = -35.9806$ \hfill \textcolor{red}{$r_{i,g,k}^{\mathrm{proc}} = -5.2327$}
    
    \item \textbf{Step 2:} First, translating the term "Übergangszeitraum" as "transitional period" makes sense in many contexts. "Transitional" is straightforward here. The phrase "der Zeitraum zwischen dem 1. Januar 1999 und dem 31. Dezember 2001" becomes "the period between 1 January 1999 and 31 December 2001." \\
    \textit{Metrics:} $\phi = -32.5788$ \hfill \textcolor{green!60!black}{$r_{i,g,k}^{\mathrm{proc}} = +3.4018$}
    
    \item \textbf{Step 3:} I should check the date format. In European-style writing, they use the ordinal, like "31. December," but in English, we use numerals without the ordinal suffix, so "31 December." Also, the years are written as 1999 and 2001. \\
    \textit{Metrics:} $\phi = -32.9279$ \hfill \textcolor{red}{$r_{i,g,k}^{\mathrm{proc}} = -0.3491$}
    
    \item \textbf{Step 4:} Putting it all together: "- 'Transitional period': the period between 1 January 1999 and 31 December 2001." \\
    \textit{Metrics:} $\phi = -30.3991$ \hfill \textcolor{green!60!black}{$r_{i,g,k}^{\mathrm{proc}} = +2.5288$}
    
    \item \textbf{Step 5:} I need to confirm if there's any other terminology here. The style is formal, so "period" is correct. No need for extra words. The original uses "der Zeitraum," which is straightforward, so "the period" works. \\
    \textit{Metrics:} $\phi = -30.0263$ \hfill \textcolor{green!60!black}{$r_{i,g,k}^{\mathrm{proc}} = +0.3728$}
    
    \item \textbf{Step 6:} I don't see any technical jargon beyond this, so the translation should be straightforward. No need for adjustments. \\
    \textit{Metrics:} $\phi = -30.7830$ \hfill \textcolor{red}{$r_{i,g,k}^{\mathrm{proc}} = -0.7567$}
    
    \item \textbf{Step 7:} Also, notice that in the original, there's a hyphen at the start, which I should keep. So the translation starts with "-". \\
    \textit{Metrics:} $\phi = -30.4701$ \hfill \textcolor{green!60!black}{$r_{i,g,k}^{\mathrm{proc}} = +0.3129$}
    
    \item \textbf{Step 8:} Double-checking the dates: January 1 to December 31. Yes, that's the correct range. \\
    \textit{Metrics:} $\phi = -30.4014$ \hfill \textcolor{green!60!black}{$r_{i,g,k}^{\mathrm{proc}} = +0.0687$}
    
    \item \textbf{Step 9:} So the final translation should be: \\
    \textit{Metrics:} $\phi = -30.7416$ \hfill \textcolor{red}{$r_{i,g,k}^{\mathrm{proc}} = -0.3401$}
    
    \item \textbf{Step 10:} - "Transitional period": the period between 1 January 1999 and 31 December 2001. \\
    \textit{Metrics:} $\phi = -29.0100$ \hfill \textcolor{green!60!black}{$r_{i,g,k}^{\mathrm{proc}} = +1.7316$}
    
    \item \textbf{Step 11:} Yes, that's concise and matches the formal style required. The colon is kept as in the original. \\
    \textit{Metrics:} $\phi = -28.6008$ \hfill \textcolor{green!60!black}{$r_{i,g,k}^{\mathrm{proc}} = +0.4091$}
    
    \item \textbf{Step 12:} I think that's all. No other nuances in the German sentence. The translation should be accurate and precise, which this seems to be. \\
    \textit{Metrics:} $\phi = -28.9584$ \hfill \textcolor{red}{$r_{i,g,k}^{\mathrm{proc}} = -0.3576$}
    
    \item \textbf{Step 13:} I don't think there's any idiomatic language here, so a direct translation works. No need for adjustments to the structure. \\
    \textit{Metrics:} $\phi = -28.8792$ \hfill \textcolor{green!60!black}{$r_{i,g,k}^{\mathrm{proc}} = +0.0792$}
    
    \item \textbf{Step 14:} So the final translation is correct, and the style is formal and concise, as required. \\
    \textit{Metrics:} $\phi = -27.7316$ \hfill \textcolor{green!60!black}{$r_{i,g,k}^{\mathrm{proc}} = +1.1476$}
\end{itemize}

\end{tcolorbox}
\caption{A detailed case study demonstrating the trajectory of process potential \(\phi\) and step-level gain \(r_{i,g,k}^{\mathrm{proc}}\) during intermediate translation reasoning. Positive values (green) indicate steps that increase the likelihood of the reference translation.}
\label{fig:case_study_pamt}
\end{figure*}

\begin{figure*}[t]
\centering
\begin{tcolorbox}[
    colback=gray!5!white,
    colframe=black!70!white,
    title=\textbf{Case Study: Trajectory of Process Potential (Terminological Rejection)},
    fonttitle=\bfseries\small,
    boxrule=0.8pt,
    arc=3mm,
    left=5pt, right=5pt, top=5pt, bottom=5pt
]
\small

\textbf{Language Pair:} German $\rightarrow$ English (de-en) \hfill
\textbf{Scores:} Format: 1.0000 \,|\, Answer: 2.0548 \,|\, Total: 3.0548 \\
\rule{\linewidth}{0.4pt}
\vspace{2pt}

\textbf{Source Input:} \\
- "Übergangszeitraum": der Zeitraum zwischen dem 1. Januar 1999 und dem 31, Dezember 2001. \\
\vspace{2pt}

\textbf{Ground Truth:} \\
- "transitional period" shall mean the period between 1 January 1999 and 31 December 2001. \\
\vspace{2pt}

\textbf{Final Answer:} \\
- "Transitional period": the period from 1 January 1999 to 31 December 2001. \\
\rule{\linewidth}{0.4pt}
\vspace{4pt}

\textbf{Step-by-Step Reasoning Analysis:} \hfill \textit{Initial} $\phi = -30.7478$ \\
\vspace{2pt}

\begin{itemize}
    \setlength{\itemsep}{2pt}
    \setlength{\parskip}{0pt}
    
    \item \textbf{Step 1:} Okay, I need to translate the German sentence "- 'Übergangszeitraum': der Zeitraum zwischen dem 1. Januar 1999 und dem 31, Dezember 2001." into English. First, let's break down the sentence. \\
    \textit{Metrics:} $\phi = -35.0311$ \hfill \textcolor{red}{$r_{i,g,k}^{\mathrm{proc}} = -4.2833$}
    
    \item \textbf{Step 2:} "Übergangszeitraum" is "transition period" or "transitional period". "den Zeitraum zwischen dem 1. Januar 1999 und dem 31. Dezember 2001" translates to "the period from 1 January 1999 to 31 December 2001". \\
    \textit{Metrics:} $\phi = -34.4548$ \hfill \textcolor{green!60!black}{$r_{i,g,k}^{\mathrm{proc}} = +0.5764$}
    
    \item \textbf{Step 3:} So the translation would be: "- 'Transition period': the period from 1 January 1999 to 31 December 2001." \\
    \textit{Metrics:} $\phi = -35.7834$ \hfill \textcolor{red}{$r_{i,g,k}^{\mathrm{proc}} = -1.3287$}
    
    \item \textbf{Step 4:} Now, I need to identify the domain. The text is about a specific time frame, so it's likely legal or governmental. The style should be formal, so using "transition period" with a colon after the term is correct. \\
    \textit{Metrics:} $\phi = -36.7196$ \hfill \textcolor{red}{$r_{i,g,k}^{\mathrm{proc}} = -0.9362$}
    
    \item \textbf{Step 5:} Check for any nuances. The German uses "bzw." (between), so the translation is accurate. Also, the dates are straightforward, no need for additional words. \\
    \textit{Metrics:} $\phi = -36.7160$ \hfill \textcolor{green!60!black}{$r_{i,g,k}^{\mathrm{proc}} = +0.0036$}
    
    \item \textbf{Step 6:} I should confirm that "Übergangszeitraum" is best as \underline{"transition period" rather than} "transitory period" or \underline{"transitional period"}. Since the original uses "Übergangszeitraum", "transition" fits well. \\
    \textit{Metrics:} $\phi = -37.5387$ \hfill \textcolor{orange!80!black}{\textbf{$r_{i,g,k}^{\mathrm{proc}} = -0.8227$}}
    
    \item \textbf{Step 7:} No other adjustments needed. The sentence structure is simple, so the translation should mirror that simplicity. \\
    \textit{Metrics:} $\phi = -36.3380$ \hfill \textcolor{green!60!black}{$r_{i,g,k}^{\mathrm{proc}} = +1.2007$}
    
    \item \textbf{Step 8:} Final translation: "- 'Transition period': the period from 1 January 1999 to 31 December 2001." \\
    \textit{Metrics:} $\phi = -36.3168$ \hfill \textcolor{green!60!black}{$r_{i,g,k}^{\mathrm{proc}} = +0.0212$}
    
    \item \textbf{Step 9:} Yes, that's concise and formal, which suits the domain style. No extra fluff, just the necessary information. \\
    \textit{Metrics:} $\phi = -35.6590$ \hfill \textcolor{green!60!black}{$r_{i,g,k}^{\mathrm{proc}} = +0.6578$}
    
    \item \textbf{Step 10:} I think that's all. The domain is probably legal or financial, but since the text is a definition, the style is formal. The translation seems to meet that requirement. \\
    \textit{Metrics:} $\phi = -35.1589$ \hfill \textcolor{green!60!black}{$r_{i,g,k}^{\mathrm{proc}} = +0.5001$}
    
    \item \textbf{Step 11:} I don't see any other issues here. The numbers and dates are correctly rendered, and the term is appropriately translated without any unnecessary complexity. \\
    \textit{Metrics:} $\phi = -34.6357$ \hfill \textcolor{green!60!black}{$r_{i,g,k}^{\mathrm{proc}} = +0.5232$}
    
    \item \textbf{Step 12:} So, the final answer is correct and the domain identification is accurate. The translation reflects the formal style required by the context. \\
    \textit{Metrics:} $\phi = -34.4831$ \hfill \textcolor{green!60!black}{$r_{i,g,k}^{\mathrm{proc}} = +0.1526$}
\end{itemize}

\end{tcolorbox}
\caption{Case study demonstrating terminological rejection. In Step 6 (\textcolor{orange!80!black}{\textbf{$r_{i,g,k}^{\mathrm{proc}} = -0.8227$}}), the model explicitly considers the correct ground-truth terminology ("\underline{transitional period}") but erroneously talks itself out of it, dismissing it in favor of "\underline{transition period}". The process potential accurately penalizes this active deviation from the reference terminology.}
\label{fig:case_study_rejection}
\end{figure*}

To elucidate how PAMT resolves the credit assignment bottleneck in reasoning-augmented MT, we contrast two diverse reasoning trajectories sampled from the exact same source sentence during the RL rollout phase (Figures \ref{fig:case_study_pamt} and \ref{fig:case_study_rejection}).
A persistent challenge in applying LRM to MDMT is terminological rejection—instances where a model successfully retrieves a correct domain constraint in its thought process but fails to execute it in the final translation. Standard sequence-level outcome rewards (e.g., metric-based reward) struggle to optimize this, as they cannot isolate where the reasoning went wrong. PAMT addresses this through dense, step-level process gains \(r_{i,g,k}^{\mathrm{proc}}\). 
In the well-aligned trajectory (Figure \ref{fig:case_study_pamt}), the policy accurately hypothesizes and commits to the target-domain term ("transitional period") early in the reasoning chain (Step 2). The process reward immediately isolates and validates this pivotal decision with a substantial gain (\(r_{i,g,k}^{\mathrm{proc}} = +3.4018\)), explicitly crediting the exact moment of domain alignment.
Conversely, the misaligned trajectory (Figure \ref{fig:case_study_rejection}) exposes a critical reasoning flaw. In Step 6, the model explicitly contemplates the ground-truth term but actively talks itself out of it, dismissing it in favor of a sub-optimal alternative ("transition period"). While an outcome-based reward would merely assign a slightly lower overall score to the final translation, PAMT's process potential explicitly isolates this incorrect decision, applying a direct step-level penalty (\(r_{i,g,k}^{\mathrm{proc}} = -0.8227\)).
Takeaway: This comparison confirms that PAMT does not merely optimize final string matching via trial and error. By directly evaluating the marginal process gain \(r_{i,g,k}^{\mathrm{proc}}\) of each intermediate step, PAMT explicitly steers the model's translation decision, ensuring that domain-faithful reasoning is reliably credited, sustained, and executed.

\begin{algorithm*}[t]
\small
\caption{RL training of \textsc{PAMT}}
\label{alg:pamt}
\DontPrintSemicolon

\SetKwInOut{KwIn}{Input}
\SetKwInOut{KwOut}{Output}

\KwIn{policy $\pi_\theta$; frozen reference model $\pi_{\mathrm{ref}}$; training set $\mathcal{D}$; metric set $\mathcal{Q}$; group size $G$; process weight $\lambda$; clip ratio $\epsilon$; KL coefficient $\beta$; number of GRPO optimization epochs $E_{\mathrm{opt}}$}
\KwOut{trained policy $\pi_\theta$}

\While{not converged}{
    $\theta_{\mathrm{old}} \leftarrow \theta$\;
    Sample a prompt minibatch $\mathcal{B}=\{(x_i,y_i^*)\}_{i=1}^{B}$ from $\mathcal{D}$\;
    Initialize rollout buffer $\mathcal{R}\leftarrow \emptyset$\;

    \tcp{Phase 1: collect rollouts and compute rewards/advantages with fixed $\pi_{\theta_{\mathrm{old}}}$}
    \For{$i \leftarrow 1$ \KwTo $B$}{
        Sample $G$ rollouts $\{o_{i,g}\}_{g=1}^{G}\sim \pi_{\theta_{\mathrm{old}}}(\cdot\mid x_i)$\;
        
        \For{$g \leftarrow 1$ \KwTo $G$}{
            Parse $o_{i,g}$ into $\texttt{<think>}z_{i,g}\texttt{</think><answer>}y_{i,g}\texttt{</answer>}$ according to Eq.~\eqref{eq:response_format}\;
            
            Split $z_{i,g}$ by double newlines into steps $[z_{i,g}^{(1)},\dots,z_{i,g}^{(K_{i,g})}]$\;
            
            Cache old token log-probabilities
            $\ell_{i,g,t}^{\mathrm{old}} \leftarrow \log \pi_{\theta_{\mathrm{old}}}(o_{i,g,t}\mid x_i,o_{i,g,<t})$
            for all response token positions $t$\;
            
            Compute format reward $r_{i,g}^{\mathrm{fmt}}$ and outcome reward $r_{i,g}^{\mathrm{out}}$ using Eq.~\eqref{eq:outcome_reward}\;
            
            Initialize $r_{i,g,t}^{\mathrm{proc}} \leftarrow 0$ for all response token positions $t$\;
            
            Compute the initial process potential $\phi_{i,g,0}$ from the empty reasoning prefix\;
            
            \For{$k \leftarrow 1$ \KwTo $K_{i,g}$}{
                Construct the prefix context $c_{i,g,k}$ by Eq.~\eqref{eq:prefix_context}\;
                
                Compute the process potential $\phi_{i,g,k}$ by Eq.~\eqref{eq:process_potential}\;
                
                Compute the step-level process gain
                $r_{i,g,k}^{\mathrm{proc}} \leftarrow \phi_{i,g,k}-\phi_{i,g,k-1}$
                by Eq.~\eqref{eq:process_gain}\;
                
                Distribute $r_{i,g,k}^{\mathrm{proc}}$ uniformly to all tokens in step $z_{i,g}^{(k)}$
                to obtain $r_{i,g,t}^{\mathrm{proc}}$ by Eq.~\eqref{eq:token_process_reward}\;
            }
            
            Compute the token-level reward sequence $\{r_{i,g,t}\}$ by Eq.~\eqref{eq:token_reward}\;
            
            Compute return-to-go $\{R_{i,g,t}\}$ by Eq.~\eqref{eq:return_to_go} and set
            $R_{i,g}^{\mathrm{traj}} \leftarrow R_{i,g,1}$\;
        }
        
        Compute group statistics $\mu_i,\sigma_i$ from
        $\{R_{i,g}^{\mathrm{traj}}\}_{g=1}^{G}$\;
        
        \For{$g \leftarrow 1$ \KwTo $G$}{
            \ForEach{response token position $t$ in rollout $o_{i,g}$}{
                Compute the token-level advantage
                \[
                A_{i,g,t} \leftarrow \frac{R_{i,g,t}-\mu_i}{\sigma_i+\epsilon}
                \]
                according to Eq.~\eqref{eq:advantage}\;
            }
            
            Add $(x_i,o_{i,g},\{\ell_{i,g,t}^{\mathrm{old}}\}_t,\{A_{i,g,t}\}_t)$ to rollout buffer $\mathcal{R}$\;
        }
    }

    \tcp{Phase 2: optimize $\pi_\theta$ on the fixed rollout buffer}
    \For{$e \leftarrow 1$ \KwTo $E_{\mathrm{opt}}$}{
        \ForEach{optimization minibatch $\widetilde{\mathcal{B}}\subset\mathcal{R}$}{
            Compute the importance ratio in Eq.~\eqref{eq:importance_ratio} using
            $\pi_\theta$ and cached $\{\ell_{i,g,t}^{\mathrm{old}}\}$\;
            
            Compute the GRPO objective on $\widetilde{\mathcal{B}}$
            by Eq.~\eqref{eq:pamt_loss}\;
            
            Take one optimizer step on $\theta$\;
        }
    }
}
\end{algorithm*}

\definecolor{samegreen}{RGB}{28,120,70}
\definecolor{newred}{RGB}{180,65,40}
\newcommand{\same}[1]{\textcolor{samegreen}{#1}}
\newcommand{\newp}[1]{\textcolor{newred}{#1}}

\section{From GRPO to Process-Aware Credit Assignment}
\label{app:grpo_to_pamt}

For clarity, we first rewrite vanilla GRPO as terminal-only token credit assignment, and then show how process reward changes the return-to-go at each token position. We use the same rollout notation as in the main text: \(i\) indexes the source sentence, \(g\) indexes the rollout sampled for that source sentence, and \(t\) indexes response tokens.

\paragraph{Vanilla GRPO.}
In vanilla GRPO, the sequence-level reward is placed only on the last valid token:
\begin{equation}
r_{i,g,t}
=
\mathbf{1}[t=t_{i,g}^{\mathrm{last}}]
\bigl(r_{i,g}^{\mathrm{fmt}}+r_{i,g}^{\mathrm{out}}\bigr),
\qquad (\lambda = 0).
\label{eq:app_vanilla_token_reward}
\end{equation}
The return-to-go is therefore
\begin{equation}
R_{i,g,t}
=
\sum_{u=t}^{T_{i,g}} r_{i,g,u}
=
r_{i,g}^{\mathrm{fmt}}+r_{i,g}^{\mathrm{out}},
\qquad
t \le t_{i,g}^{\mathrm{last}} .
\label{eq:app_vanilla_rtg}
\end{equation}
Thus, reward-to-go propagates the same terminal reward to all previous tokens, and the trajectory return is \(R_{i,g}^{\mathrm{traj}}=R_{i,g,1}\). For the \(G\) rollouts sampled from the same source sentence \(x_i\), we compute
\begin{equation}
\mu_i
=
\frac{1}{G}\sum_{g=1}^{G} R_{i,g}^{\mathrm{traj}} ,
\label{eq:app_group_mean}
\end{equation}
\begin{equation}
\sigma_i
=
\sqrt{
\frac{1}{G}\sum_{g=1}^{G}
\left(R_{i,g}^{\mathrm{traj}}-\mu_i\right)^2 } .
\label{eq:app_group_std}
\end{equation}
Hence every valid token in the same rollout shares the same advantage:
\begin{equation}
A_{i,g,t}
=
\frac{R_{i,g,t}-\mu_i}{\sigma_i+\epsilon}.
\label{eq:app_vanilla_adv}
\end{equation}

\paragraph{Step-level process reward.}
To obtain process-level supervision, we first define a zero-step prefix:
\begin{equation}
c_{i,g,0}
=
[x_i,\texttt{<think>}\texttt{</think>},\texttt{<answer>}].
\label{eq:app_zero_prefix}
\end{equation}
For \(k=1,\dots,K_{i,g}\), let
\begin{equation}
c_{i,g,k}
=
[
x_i,\texttt{<think>},z_{i,g,\le k},
\texttt{</think>},\texttt{<answer>}
].
\label{eq:app_prefix}
\end{equation}
We define the process potential of the \(k\)-step prefix as
\begin{equation}
\phi_{i,g,k}
=
\sum_{m=1}^{|y_i^*|}
\log \pi_{\mathrm{ref}}
\bigl(
y_{i,m}^* \mid c_{i,g,k}, y_{i,<m}^*
\bigr).
\label{eq:app_potential}
\end{equation}
The step-level process gain is then
\begin{equation}
r_{i,g,k}^{\mathrm{proc}}
=
\phi_{i,g,k}-\phi_{i,g,k-1},
\qquad
k=1,\dots,K_{i,g}.
\label{eq:app_step_gain}
\end{equation}
A positive \(r_{i,g,k}^{\mathrm{proc}}\) means that adding step \(z_{i,g}^{(k)}\) makes the reference translation more predictable under \(\pi_{\mathrm{ref}}\).

To optimize at the token level, we distribute each step gain uniformly over the tokens in that step:
\begin{equation}
r_{i,g,t}^{\mathrm{proc}}
=
\begin{cases}
\dfrac{r_{i,g,k}^{\mathrm{proc}}}{|z_{i,g}^{(k)}|},
& \text{if token } t \text{ belongs to step } z_{i,g}^{(k)}, \\[8pt]
0, & \text{otherwise.}
\end{cases}
\label{eq:app_token_proc}
\end{equation}
This preserves the total gain of each reasoning step while avoiding a length bias toward longer steps.

\paragraph{Unified token reward.}
After adding process reward, the token-level reward becomes
\begin{equation}
r_{i,g,t}
=
\same{\mathbf{1}[t=t_{i,g}^{\mathrm{last}}]
\bigl(r_{i,g}^{\mathrm{fmt}}+r_{i,g}^{\mathrm{out}}\bigr)}
+
\newp{\lambda\, r_{i,g,t}^{\mathrm{proc}}}.
\label{eq:app_token_reward}
\end{equation}
Compared with vanilla GRPO, the \same{green term} is unchanged, while the \newp{red term} is the newly introduced dense process supervision.

The corresponding return-to-go is
\begin{equation}
\begin{aligned}
R_{i,g,t}
&=
\sum_{u=t}^{T_{i,g}} r_{i,g,u} \\
&=
\same{r_{i,g}^{\mathrm{fmt}}+r_{i,g}^{\mathrm{out}}}
+
\newp{\lambda \sum_{u=t}^{T_{i,g}} r_{i,g,u}^{\mathrm{proc}}}.
\end{aligned}
\label{eq:app_rtg}
\end{equation}
Equation~\eqref{eq:app_rtg} reduces to Eq.~\eqref{eq:app_vanilla_rtg} when \(\lambda=0\), i.e., standard GRPO.

More explicitly, suppose token \(t\) is the \(\ell\)-th token in reasoning step \(z_{i,g}^{(k)}\), where \(\ell \in \{1,\dots,|z_{i,g}^{(k)}|\}\). Then
\begin{equation}
\begin{aligned}
R_{i,g,t}
=
r_{i,g}^{\mathrm{fmt}}+r_{i,g}^{\mathrm{out}}
+ \lambda \Biggl(
&\frac{|z_{i,g}^{(k)}|-\ell+1}{|z_{i,g}^{(k)}|}
r_{i,g,k}^{\mathrm{proc}} \\
&+ \sum_{j=k+1}^{K_{i,g}} r_{i,g,j}^{\mathrm{proc}}
\Biggr).
\end{aligned}
\label{eq:app_rtg_explicit}
\end{equation}
Hence, different reasoning tokens aggregate different suffixes of future process reward. Tokens whose remaining reasoning suffix is more helpful receive larger returns, while tokens followed by harmful steps receive smaller returns. In contrast, for tokens in the \texttt{<answer>} span, all process rewards lie in the past, so
\begin{equation}
R_{i,g,t}
=
r_{i,g}^{\mathrm{fmt}}+r_{i,g}^{\mathrm{out}},
\qquad
t \in \texttt{<answer>} \text{ span}.
\label{eq:app_answer_rtg}
\end{equation}

The trajectory return now becomes
\begin{equation}
R_{i,g}^{\mathrm{traj}}
=
R_{i,g,1}
=
r_{i,g}^{\mathrm{fmt}} + r_{i,g}^{\mathrm{out}}
+ \lambda \sum_{k=1}^{K_{i,g}} r_{i,g,k}^{\mathrm{proc}}.
\label{eq:app_traj}
\end{equation}

\paragraph{Advantage normalization.}
As in GRPO, we normalize the token return using group statistics computed from the updated trajectory returns:
\begin{equation}
\mu_i
=
\frac{1}{G}\sum_{g=1}^{G} R_{i,g}^{\mathrm{traj}} ,
\label{eq:app_group_mean_updated}
\end{equation}
\begin{equation}
\sigma_i
=
\sqrt{
\frac{1}{G}\sum_{g=1}^{G}
\left(R_{i,g}^{\mathrm{traj}}-\mu_i\right)^2 } .
\label{eq:app_group_std_updated}
\end{equation}
The token-level advantage is then
\begin{equation}
A_{i,g,t}
=
\frac{R_{i,g,t}-\mu_i}{\sigma_i+\epsilon}.
\label{eq:app_adv}
\end{equation}
Unlike vanilla GRPO, \(A_{i,g,t}\) now depends on the token position through the remaining future process reward.

\paragraph{Optimization.}
The importance ratio is
\begin{equation}
\rho_{i,g,t}(\theta)
=
\frac{
\pi_\theta(o_{i,g,t}\mid x_i,o_{i,g,<t})
}{
\pi_{\theta_{\mathrm{old}}}(o_{i,g,t}\mid x_i,o_{i,g,<t})
}.
\label{eq:app_ratio}
\end{equation}
We further define the clipped ratio
\begin{equation}
\bar{\rho}_{i,g,t}(\theta)
=
\operatorname{clip}
\bigl(
\rho_{i,g,t}(\theta),\,1-\epsilon,\,1+\epsilon
\bigr).
\label{eq:app_clipped_ratio}
\end{equation}
and the clipped token objective
\begin{equation}
\ell_{i,g,t}(\theta)
=
\min
\Bigl(
\rho_{i,g,t}(\theta)A_{i,g,t},
\bar{\rho}_{i,g,t}(\theta)A_{i,g,t}
\Bigr).
\label{eq:app_token_obj}
\end{equation}
The GRPO-style clipped loss is
\begin{equation}
\mathcal{L}_{\mathrm{clip}}(\theta)
=
-\frac{1}{B G}\sum_{i=1}^{B}\sum_{g=1}^{G}
\frac{1}{T_{i,g}}
\sum_{t=1}^{T_{i,g}}
\ell_{i,g,t}(\theta),
\label{eq:app_clip_loss}
\end{equation}
and the full objective is
\begin{equation}
\mathcal{L}(\theta)
=
\mathcal{L}_{\mathrm{clip}}(\theta)
+
\beta\,\mathrm{KL}
\bigl(
\pi_\theta \,\|\, \pi_{\mathrm{ref}}
\bigr).
\label{eq:app_full_loss}
\end{equation}

In this view, vanilla GRPO is simply the special case \(\lambda=0\), where reward-to-go propagates the same terminal reward to all previous tokens. PAMT keeps the same GRPO optimization backbone, but changes the token return by adding dense process rewards, making token credit assignment position-dependent.

\paragraph{Toy example.}
Consider one rollout with three reasoning steps. Suppose their step-level gains are
\[
\bigl(r_{i,g,k}^{\mathrm{proc}}\bigr)_{k=1}^{3}
=
(0.4,\,-0.3,\,0.2),
\]
and their lengths are
\[
(|z_{i,g}^{(1)}|,\,|z_{i,g}^{(2)}|,\,|z_{i,g}^{(3)}|)
=
(2,\,1,\,1).
\]
Intuitively, the first step is helpful, the second step introduces process drift, and the third step partially corrects it.

By distributing each step gain uniformly over its tokens, the token-level process rewards become
\[
\bigl(r_{i,g,t}^{\mathrm{proc}}\bigr)_{t=1}^{4}
=
(0.2,\,0.2,\,-0.3,\,0.2).
\]
Suppose \(r_{i,g}^{\mathrm{fmt}}+r_{i,g}^{\mathrm{out}}=0.8\) and \(\lambda=1\). Then the return-to-go on the four reasoning tokens is
\[
(R_{i,g,1},\,R_{i,g,2},\,R_{i,g,3},\,R_{i,g,4})
=
(1.1,\,0.9,\,0.7,\,1.0),
\]
because each token receives the sequence-level reward plus the suffix sum of the remaining process rewards. For example,
\[
R_{i,g,2}
=
0.8 + (0.2 - 0.3 + 0.2)
=
0.9,
\]
while
\[
R_{i,g,3}
=
0.8 + (-0.3 + 0.2)
=
0.7.
\]
For tokens in the \texttt{<answer>} span, all process rewards already lie in the past, so
\[
R_{i,g,t}=0.8.
\]

Under vanilla GRPO, all of these positions would instead receive the same return \(0.8\). This example shows that process-aware credit assignment is not simply ``earlier is better''; rather, a token receives a higher return when the remaining reasoning suffix is more helpful for producing the reference translation.

\end{document}